\PassOptionsToPackage{table,dvipsnames}{xcolor}
\documentclass{article}
\usepackage{iclr2027_conference,times}

\usepackage{tabularx}
\usepackage{array}
\usepackage{microtype}
\usepackage{graphicx}
\usepackage{subcaption}
\usepackage{booktabs}
\usepackage[table,dvipsnames]{xcolor}
\usepackage{pifont}
\usepackage{url}
\usepackage{amsmath}
\usepackage{amssymb}
\usepackage{mathtools}
\usepackage{amsthm}
\usepackage{algorithm}
\usepackage{algorithmic}
\usepackage{enumitem}
\usepackage{tcolorbox}
\tcbuselibrary{breakable}
\usepackage{hyperref}
\hypersetup{
  hidelinks,
  pdftitle={CB-Orchestrator: Adaptive Workflow Optimization for LLM Agents via Contextual Bandits},
  pdfauthor={Jiahang Sun, Zhiwei Shang, Zhipiao Liu, Hongwei Yang, Guoqing Xie, Shuang Qiu, Zhongxiang Dai}
}

\DeclareMathOperator*{\argmax}{arg\,max}

\theoremstyle{plain}
\newtheorem{theorem}{Theorem}[section]

\theoremstyle{definition}

\newtheorem{assumption}[theorem]{Assumption}
\theoremstyle{remark}

\title{CB-Orchestrator: Adaptive Workflow Optimization for LLM Agents via Contextual Bandits}
\author{Jiahang Sun$^{1}$,
Zhiwei Shang$^{1}$,
Zhipiao Liu$^{2}$,
Hongwei Yang$^{2}$,
Guoqing Xie$^{2}$,\\
\textbf{Shuang Qiu}$^{3}$,
\textbf{Zhongxiang Dai}$^{1}$%
\thanks{Corresponding author. Correspondence to daizhongxiang@cuhk.edu.cn.}
\\[1ex]
$^{1}$The Chinese University of Hong Kong, Shenzhen\\
$^{2}$Huawei Technologies \quad $^{3}$City University of Hong Kong
}

\iclrfinalcopy
\date{}

\begin{document}
\maketitle

\begin{abstract}
Large Language Model (LLM) agents have demonstrated remarkable capabilities in tackling complex tasks through agentic workflows. However, manually designing these workflows is labor-intensive and lacks flexibility to adapt to diverse queries. While automated workflow optimization is a promising direction, existing methods often incur expensive API costs during training-phase feedback collection. Furthermore, they fail to adaptively reuse high-quality workflows, which creates a bottleneck for both performance and cost-effectiveness. To address these limitations, we propose \textbf{CB-Orchestrator}, an adaptive framework that \textbf{decouples workflow generation from selection}. In the first stage, we construct a diverse pool of candidate workflows via evolutionary search. In the second stage, we formulate workflow selection as a \textbf{Contextual Bandit} problem, which enables sample-efficient learning by balancing exploration and exploitation, thereby significantly reducing the feedback required for training. During inference, the model adaptively selects the optimal workflow tailored to each query. Evaluations across five diverse benchmarks demonstrate that CB-Orchestrator consistently outperforms all baselines. Notably, on the MATH dataset, CB-Orchestrator achieves better performance than strong recent baselines while reducing training-phase API token overhead by no less than 40.55\% and total end-to-end training (including workflow pool generation) token overhead by no less than 29.73\%.

\end{abstract}

\section{Introduction}
\label{sec:introduction}

Large Language Model (LLM) agents have demonstrated remarkable capabilities in handling complex tasks through \emph{agentic workflows}~\citep{autogen, metagpt}. An \emph{agentic workflow} is a structured program that orchestrates a sequence of LLM calls and tool invocations to solve a query~\citep{ADAS, AFlow}. However, manually designing such workflows is labor-intensive and often fails to adapt to diverse queries. This has motivated growing interest in automated workflow optimization.

\begin{figure}[htbp]
    \centering
    \includegraphics[width=0.99\linewidth]{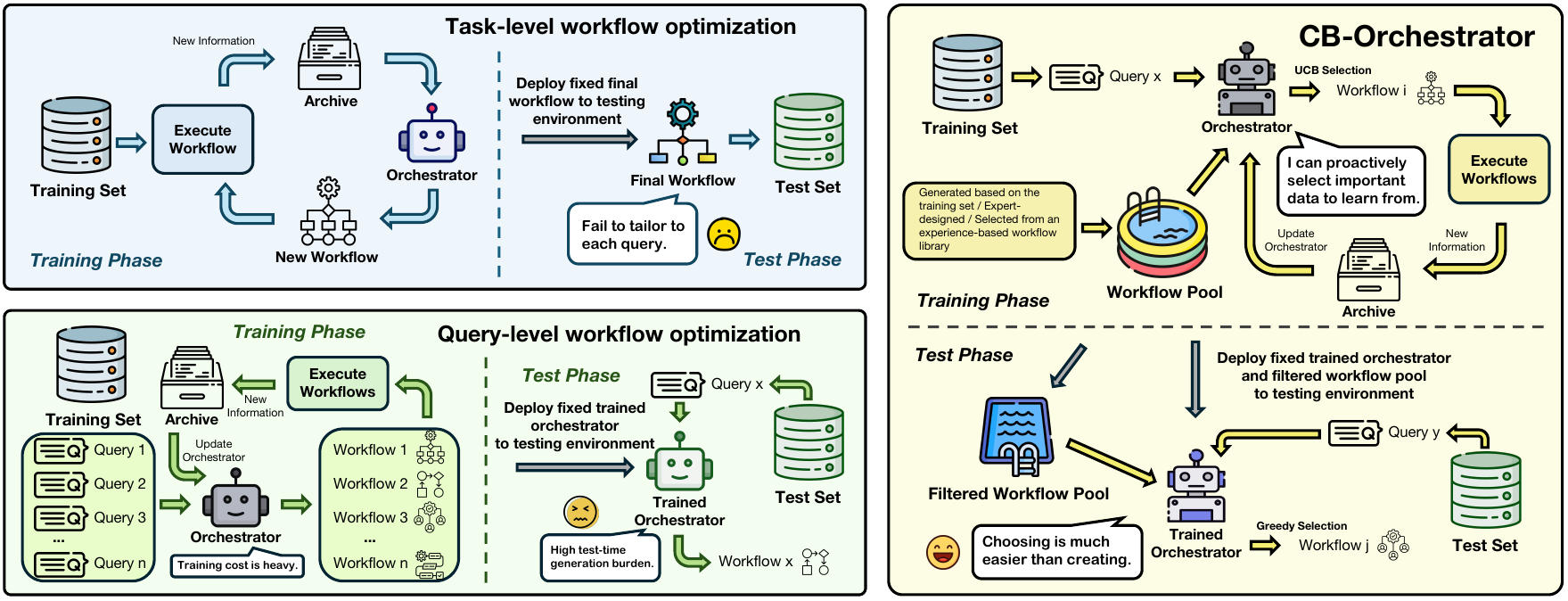}
    \caption{CB-Orchestrator vs.\ prior workflow-optimization paradigms.}
    \label{fig:1}
\end{figure}

As illustrated in Figure~\ref{fig:1}, current automatic workflow optimization largely falls into two paradigms: (i) \emph{task-level optimization}, which implements a universal fixed workflow for a task type~\citep{AutoFlow, ADAS, AFlow}; and (ii) \emph{query-level optimization}, which generates a bespoke workflow for each incoming query~\citep{zhang2025evoflow, zhu2025adaptflow, zhang2025metaagent}. However, these paradigms present a fundamental dilemma. Task-level approaches suffer from rigidity and fail to tailor solutions to diverse queries. In contrast, query-level approaches treat every query as a new generation problem. This significantly harder learning objective creates a bottleneck, inflating both the sample complexity needed for training-phase feedback collection and the computational overhead required for inference. Furthermore, neither paradigm effectively reuses high-quality workflows: learned solutions are either too generic or discarded after one-off use, preventing the system from benefiting from accumulated knowledge.

Beyond efficiency bottlenecks, a fundamental issue is that \emph{workflow effectiveness is inherently query-dependent}. A single fixed workflow optimized at the task level can be unnecessarily heavy for simple instances yet insufficient for complex ones (e.g., arithmetic vs. multi-step verification in math reasoning; one-shot utility functions vs. iterative testing in code generation). While query-level methods address this heterogeneity, they often conflate \emph{adaptation} with \emph{generation}: generating a new workflow from scratch is a difficult program-synthesis problem with a vast search space, whereas selecting a suitable strategy from a reusable set is a much simpler decision problem. This motivates our key rationale: \emph{choosing is much easier than creating}. We therefore aim to \emph{reuse} a pool of strong workflows and spend the learning budget on \emph{adaptive selection}.

To this end, we propose \textbf{CB-Orchestrator}, which \textbf{decouples workflow pool construction from adaptive selection} (Figure~\ref{fig:1}). CB-Orchestrator proceeds in two stages: Stage~1 constructs a diverse workflow pool automatically from the training set via execution-guided evolutionary search, and Stage~2 learns a query-conditioned selecting policy by formulating workflow selection as a \emph{contextual bandit}. The bandit view enables a principled exploration--exploitation trade-off, allowing the selector to learn sample-efficiently.

\textbf{Contributions.}
We make the following key contributions:
\begin{enumerate}[leftmargin=*,label=(\arabic*)]
    \item \textbf{Problem Setting:} We formalize \emph{adaptive workflow reuse} for LLM agents as learning a query-conditioned routing policy over a reusable workflow pool, as an alternative to task-level fixed-workflow deployment and query-level per-query workflow generation (Figure~\ref{fig:1}).
    \item \textbf{CB-Orchestrator:} We propose a two-stage framework that (i) constructs a diverse workflow pool \emph{automatically from the training set} via execution-guided evolutionary search, and (ii) learns a \emph{contextual-bandit} selector to adaptively choose workflows, enabling sample-efficient training by reducing the number of workflow evaluations needed to learn a strong policy.
    \item \textbf{Results:} Across five benchmarks, CB-Orchestrator consistently improves over strong baselines. On the MATH dataset, it reduces training-phase token overhead by $\ge 40.55\%$ and end-to-end training overhead (including workflow pool generation) by $\ge 29.73\%$.
\end{enumerate}

\section{Related Work}
\label{sec:related_work}

\subsection{Agentic Workflows and Autonomous Agents}
LLM applications can be organized into two paradigms: \emph{agentic workflows} and \emph{autonomous agents}~\citep{AFlow}. Agentic workflows decompose a task into a structured sequence of model invocations and tool calls, whereas autonomous agents follow a more flexible, stateful control policy that dynamically decides the next action based on intermediate outcomes and environment feedback. Agentic workflows generalize single-instance reasoning strategies (e.g., Chain-of-Thought (CoT) \citep{CoT} and self-consistency \citep{Self-consistency}) into closed-loop architectures. Representative designs such as Self-Refine \citep{Self-refine} and Reflexion \citep{Reflexion} incorporate execution feedback or self-evaluation signals to iteratively update intermediate solutions, enabling calibration over multiple turns rather than relying on a one-shot generation.

Autonomous agents maintain an explicit \emph{state} (e.g., memory, plans, and tool results) and use a policy-like controller to select actions online from intermediate outcomes and environment feedback. Such designs are realized in agent frameworks (e.g., CAMEL \citep{camel}, AutoGen \citep{autogen}, and MetaGPT \citep{metagpt}), but are often labor-intensive to specify (prompts, roles, tools, and interaction topologies). 

\subsection{Automated Workflow Optimization}
Recent progress has shifted from \emph{manually designed} workflows to \emph{automatically optimized} agentic systems. 
At the \emph{task level}, the goal is to learn a \emph{query-invariant} workflow for a given task/benchmark by optimizing a dataset-level objective.
Representative examples include AutoFlow \citep{AutoFlow}, ADAS \citep{ADAS}, AgentSquare \citep{AgentSquare}, and AFlow \citep{AFlow}, which explore different search paradigms (e.g., heuristic search, modular search, and MCTS-style exploration). 
Beyond workflow discovery, several works broaden the optimization toolbox for multi-agent collaboration and evolution, including MASS \citep{zhou2025multi}, OMAC \citep{li2025omac}, SEW \citep{liu2025sew}, and safety-aware evolutionary programming in MermaidFlow \citep{zheng2025mermaidflow}. 
Swarm-intelligence-based generation (SwarmAgentic) \citep{zhang2025swarmagentic} further pushes toward automated end-to-end system construction.

In addition to searching over entire workflows, some approaches optimize \emph{inter-agent communication structures} and \emph{topologies} directly, treating MAS as optimizable graphs.
GPTSwarm \citep{GPTSwarm} models language agents as graph-structured systems, while G-Designer \citep{zhang2024gdesigner} explicitly architects multi-agent communication via graph learning. More broadly, supernet-style architecture optimization has been explored in MaAS \citep{MaAS}, which learns an agentic supernet (a distribution over architecture) instead of a single fixed workflow.

At the \emph{query level}, systems plan or adapt workflows conditioned on each input \citep{zhang2025evoflow, zhu2025adaptflow, zhang2025metaagent}. 
Researchers train agents (or meta-agents) to construct or orchestrate MAS automatically, such as MAS-GPT \citep{ye2025masgpt}, Weak-for-Strong \citep{nie2025weakforstrong} and ScoreFlow \citep{wang2025scoreflow}. 
Complementary efforts study dynamic agent networks \citep{liu2024dynamic}, query-level meta-agent reinforcement \citep{gao2025flowreasoner}, robust workflow generation \citep{xu2025robustflow}, modularized workflow automation \citep{niu2025flow}, and self-optimizing hierarchical workflows \citep{ho2025polymath}. 
Researchers also explore efficient refinement signals, including difficulty-aware orchestration \citep{su2025difficulty} and failure-driven refinement \citep{zhang2025failure}.
In contrast to prior work that primarily (i) optimizes a single dataset-level workflow or (ii) directly generates/refines workflows per query, our setting focuses on \emph{adaptive reuse and selection} of workflows.

\section{Methodology}
\label{sec:methodology}

We present \textbf{CB-Orchestrator}, which decouples workflow pool construction from query-level selection. We first formalize the problem, then describe pool construction, contextual bandit training, and inference.

\subsection{Workflow Representation and Problem Formulation}
\label{sec:problem}

We represent workflows as executable Python classes with a standardized async callable interface. Following AFlow~\citep{AFlow}, we define a library of reusable \emph{operators} as atomic building blocks that encapsulate common LLM interaction patterns (see Appendix~\ref{app:operator_library} for details). Each workflow is composed of multiple operator instances that orchestrate their execution to solve problems.

Consider a set of queries $\mathcal{Q} = \{q_1, q_2, \ldots, q_n\}$, where each query $q_i$ has an associated ground-truth answer $a_i$. We partition $\mathcal{Q}$ into two disjoint subsets: $\mathcal{Q}_{\text{train}}$ for workflow pool construction (Stage 1) and bandit policy training (Stage 2), and $\mathcal{Q}_{\text{test}}$ for final evaluation (Stage 3). An \emph{agentic workflow} $w$ is a structured program that orchestrates LLM calls and tool invocations to produce an answer $\hat{a} = w(q)$ for a query $q$.

We formulate \emph{workflow optimization} as finding, for each query $q$, a workflow $w_q$ that maximizes the expected reward:
\begin{equation}
    w_q^* = \argmax_{w} \mathbb{E}\left[ r(q, w) \right]
\end{equation}
where $r(q, w) \in [0, 1]$ measures the quality of the answer produced by workflow $w$ for query $q$. In principle, this requires generating a customized workflow for every query, which is expensive.

CB-Orchestrator instead constructs a reusable pool of operator compositions $\mathcal{W} = \{w_1, \ldots, w_K\}$ and learns a query-level selection policy $\pi: \mathcal{Q} \rightarrow \mathcal{W}$ to choose a workflow from the pool.

\subsection{Stage 1: Workflow Pool Construction}
\label{sec:workflow_generation}

The first stage constructs a diverse workflow pool $\mathcal{W}$ as the arm set for bandit selection. Construction targets \textbf{validity} ($\forall w \in \mathcal{W},\, \exists q \in \mathcal{Q}_{\text{train}}: r(q,w) \geq \tau$) and \textbf{completeness} ($\forall q \in \mathcal{Q}_{\text{train}},\, \exists w \in \mathcal{W}: r(q,w) \geq \tau$). Completeness is a design goal; measured pool coverage is reported in Appendix Table~\ref{tab:pool_coverage}.

The pool can contain human-designed, domain-validated, or automatically discovered workflows; for fair comparison, we use the latter. Given seed queries $\mathcal{Q}_{\text{seed}} \subseteq \mathcal{Q}_{\text{train}}$, we apply three steps (hyperparameters in Appendix~\ref{app:implementation}; prompts in Appendix~\ref{app:prompts}):

\begin{itemize}[leftmargin=1.5em]
    \item \textbf{Initial Generation.} For each $q \in \mathcal{Q}_{\text{seed}}$, prompt an LLM to generate an initial workflow $w_0^{(q)} = \textsc{Generate}(q, \mathcal{O}, \mathcal{E})$, where $\mathcal{O}$ is the operator library and $\mathcal{E}$ contains examples. Successful workflows ($r(q, w) \geq \tau$) are added to $\mathcal{W}$.

    \item \textbf{Execution-Guided Evolution.} If the initial workflow fails, evolve it using execution feedback:
    \begin{equation}
        w_{j+1}^{(q)} = \textsc{Evolve}(w_j^{(q)}, q, \text{trace}_j, \text{error}_j)
    \end{equation}
    where $\text{trace}_j$ captures operator-level execution traces and $\text{error}_j$ contains error messages. Repeat until success or reaching the maximum evolution rounds.

    \item \textbf{Cross-Query Transfer.} For queries where evolution fails, retrieve similar successful queries using embedding similarity:
    \begin{equation}
        \mathcal{N}_k(q) = \argmax_{(q', w') \in \mathcal{S}}^{(k)} \cos(\phi(q), \phi(q'))
    \end{equation}
    where $\mathcal{S}$ is the set of successful (query, workflow) pairs. Test if retrieved workflows directly solve $q$; otherwise, evolve them further.
\end{itemize}

\subsection{Stage 2: Contextual Bandit Selection}
\label{sec:bandit_selection}

Stage 2 treats the query embedding $\phi(q) \in \mathbb{R}^d$ as context, workflows $w \in \mathcal{W}$ represented by $\phi(w) \in \mathbb{R}^d$ as arms, and answer quality $r \in [0,1]$ as reward (Section~\ref{sec:feature_rep}). Our Hybrid Neural-Linear UCB selector combines neural reward prediction with a linear exploration bonus to balance performance and sample efficiency under uncertain outcomes.

\subsubsection{Bandit Formulation and Assumptions}

At each round $t$, the learner observes $q_t \sim \mathcal{Q}_{\text{train}}$ and its context $\phi(q_t)$, selects an arm $w_t \in \mathcal{W}$, and receives a bounded reward $r_t \in [0,1]$.

We make the following assumptions for contextual bandits:
\begin{assumption}[Semantic Embedding]
\label{assumption:semantic}
The embedding function $\phi$ preserves semantic similarity: queries requiring similar reasoning patterns have similar embeddings, and workflows implementing similar problem-solving strategies have similar embeddings. Formally, if $q_1$ and $q_2$ require similar skills to solve, then $\|\phi(q_1) - \phi(q_2)\|$ is small; if $w_1$ and $w_2$ implement similar operator compositions, then $\|\phi(w_1) - \phi(w_2)\|$ is small.
\end{assumption}

\begin{assumption}[Reward Smoothness]
\label{assumption:smoothness}
The expected reward function is Lipschitz continuous in the embedding space: semantically similar (query, workflow) pairs yield similar expected rewards. Formally, there exists $L > 0$ such that $|\mathbb{E}[r|q_1, w_1] - \mathbb{E}[r|q_2, w_2]| \leq L \|x(q_1, w_1) - x(q_2, w_2)\|$.
\end{assumption}

\begin{assumption}[Realizability]
\label{assumption:realizability}
There exists a function $f^*: \mathbb{R}^{2d} \rightarrow [0, 1]$ such that $\mathbb{E}[r | q, w] = f^*(x(q, w))$, where $x(q, w) = [\phi(q); \phi(w)] \in \mathbb{R}^{2d}$ is the joint feature representation (see Section~\ref{sec:feature_rep}).
\end{assumption}

Empirical validation of these assumptions is provided in Appendix~\ref{app:assumption_validation}.

\subsubsection{Feature Representation}
\label{sec:feature_rep}

For a (query, workflow) pair $(q, w)$, we construct a joint feature representation by concatenating their $L_2$-normalized embeddings:
\begin{equation}
    x(q, w) = [\phi(q); \phi(w)] \in \mathbb{R}^{2d}, \quad \phi(\cdot) = \frac{\text{Embed}(\cdot)}{\|\text{Embed}(\cdot)\|_2}
\end{equation}
We use a single frozen text embedding model to encode both query text and workflow source code. The workflow code serves as a semantic summary of the problem-solving strategy, capturing operator sequences and composition structure. The concatenated representation $x(q, w)$ encodes the compatibility between query requirements and workflow capabilities. Implementation details (e.g., context length handling) are provided in Appendix~\ref{app:implementation}.

\subsubsection{Hybrid Neural-Linear Architecture}

\textbf{Neural Network Component.}
We employ a 3-layer MLP with progressive dimensionality reduction to predict rewards:
\begin{equation}
    f_\theta(x) = \sigma(\text{MLP}(x)) \in [0, 1]
\end{equation}
where $\sigma$ is the sigmoid output. Architecture details are provided in Appendix~\ref{app:implementation}.

\textbf{Linear UCB Component.}
Following LinUCB~\citep{LinUCB}, we maintain a precision matrix $A = \lambda I + \sum_{t=1}^{T} x_t x_t^\top$ and compute uncertainty as:
\begin{equation}
    u(x) = \sqrt{x^\top A^{-1} x}
\end{equation}
In our hybrid design, $u(x)$ primarily measures \emph{feature-space novelty/coverage} rather than directly estimating $\mathrm{Var}[r\mid x]$; under the smoothness assumption, it serves as a practical proxy for exploration. This enables closed-form uncertainty quantification, making exploration behavior predictable via the parameter $\alpha$. Computational details are in Appendix~\ref{app:theory}.

\textbf{Hybrid UCB Score.}
The final selection score combines exploitation (neural network prediction) with exploration (linear uncertainty):
\begin{equation}
    \text{UCB}(q, w) = f_\theta(x(q, w)) + \alpha \cdot u(x(q, w))
    \label{eq:ucb}
\end{equation}
where $\alpha > 0$ controls the exploration-exploitation trade-off. During training, we select the eligible workflow with the highest score (Algorithm~\ref{alg:training}).

\begin{algorithm}[t]
\caption{CB-Orchestrator Training}
\label{alg:training}
\begin{algorithmic}[1]
\REQUIRE Workflow pool $\mathcal{W}$, training queries $\mathcal{Q}_{\text{train}}$, exploration rounds $T_{\text{explore}}$, total rounds $T$, exploration weight $\alpha$, regularization $\lambda$, evaluation batch size $B$, training epochs $E$
\STATE Initialize $f_\theta$, $A \leftarrow \lambda I$, $\mathcal{D} \leftarrow \emptyset$, $\mathcal{H} \leftarrow \emptyset$
\FOR{batch $b = 1$ to $\lceil T / B \rceil$}
    \STATE $\mathcal{P} \leftarrow \emptyset$ \hfill \textit{current batch of pairs}
    \FOR{$t = (b-1)B+1$ to $\min(bB,T)$}
        \STATE Sample $q \sim \mathcal{Q}_{\text{train}}$; $\mathcal{W}_q \leftarrow \{v \in \mathcal{W} : (q,v) \notin \mathcal{H}\}$
        \STATE \textbf{if} $t \leq T_{\text{explore}}$ \textbf{then} sample $w \sim \text{Uniform}(\mathcal{W}_q)$
        \STATE \textbf{else} set $w \leftarrow \argmax_{v \in \mathcal{W}_q} \text{UCB}(q,v)$
        \STATE $\mathcal{P} \leftarrow \mathcal{P} \cup \{(q,w)\}$; $\mathcal{H} \leftarrow \mathcal{H} \cup \{(q,w)\}$
    \ENDFOR
    \STATE $\mathcal{R} \leftarrow \text{ParallelExec}(\mathcal{P})$ \hfill \textit{returns $(q,w,r)$ triples}
    \FOR{$(q,w,r) \in \mathcal{R}$}
        \STATE $x \leftarrow [\phi(q); \phi(w)]$; $A \leftarrow A + x x^\top$; $\mathcal{D} \leftarrow \mathcal{D} \cup \{(x, r)\}$
    \ENDFOR
    \STATE Reset $f_\theta$; train on $\mathcal{D}$ for $E$ epochs
\ENDFOR
\STATE \textbf{return} $(f_\theta, A)$
\end{algorithmic}
\end{algorithm}

\subsubsection{Training Algorithm}

Algorithm~\ref{alg:training} begins with uniform exploration for $T_{\text{explore}}$ rounds, then selects workflows by Equation~\ref{eq:ucb}; both phases avoid repeated (query, workflow) evaluations. Workflows execute in parallel batches to reduce wall-clock time. After each batch, rewards update $A$ and the accumulated training buffer $\mathcal{D}$, and the reward predictor is reset and retrained for $E$ epochs. Implementation details and high-noise extensions appear in Appendix~\ref{app:implementation}.

\subsection{Stage 3: Inference with Adaptive Selection}
\label{sec:inference}

Given a test query $q \in \mathcal{Q}_{\text{test}}$, CB-Orchestrator executes the workflow with the highest predicted reward:
\begin{equation}
    w^* = \argmax_{w \in \mathcal{W}} f_\theta(x(q, w))
\end{equation}
Inference uses no exploration bonus. For robustness, Wilson score confidence intervals filter workflows with insufficient trials or low success rates (Appendix~\ref{app:implementation}). An optional online learning mode adapts the model under distribution shifts and expands the pool for queries outside its coverage (Appendix~\ref{app:online_learning}).

\section{Experiments}
\label{sec:experiments}

\subsection{Experimental Setup}
\label{subsec:setup}
\textbf{Datasets.} We conduct experiments on five diverse benchmarks: GSM8K \citep{GSM8K} and MATH \citep{MATH} for math reasoning, HumanEval \citep{HumanEval} and MBPP \citep{MBPP} for code generation, and the comprehensive GAIA \citep{GAIA} dataset for evaluation on real-world tasks requiring reasoning, multi-modality, and tool-use proficiency. For GSM8K, MATH and GAIA, we report the Solve Rate (\%); for HumanEval and MBPP, we report pass@1. Detailed statistics regarding the datasets are presented in Appendix~\ref{app:implementation}.

\textbf{Baselines.} We compare our proposed CB-Orchestrator against a wide range of competitive baselines: (1) Basic Prompting methods, including Vanilla zero-shot prompting, Chain-of-Thought (CoT) \mbox{\citep{CoT}}, and CoT Self-Consistency (CoT SC) \citep{Self-consistency} which enhance robustness through majority voting. (2) Fixed Multi-Agent Systems, which employ static collaboration structures, including LLM-Debate \citep{LLMDebate}, LLM-Blender \citep{LLMBlender}, DyLAN \citep{liu2024dynamic}, AgentVerse \citep{AgentVerse}, TapeAgent \citep{TapeAgent} and Sibyl~\citep{Sibyl}. (3) Autonomous Multi-Agent Systems, which automatically generate agent systems. This group includes AutoAgents \citep{AutoAgents}, GPTSwarm \citep{GPTSwarm} that models agents as optimizable graphs, AgentSquare \citep{AgentSquare} for modular search, and recent high-performing frameworks AFlow \citep{AFlow} and MaAS \citep{MaAS}.

\textbf{Implementation Details.} We employ GPT-4o-mini-0718 \citep{4o-mini} as the backbone LLM across all stages for both our algorithm and baselines, with Llama-3.1-70B \citep{grattafiori2024llama} used for cross-model generalization analysis. We employ the Qwen3-4B embedding model \citep{qwen3embedding}. For Stage~1, workflow generation uses temperature 1.0 for diversity, while all other LLM calls use temperature 0 for determinism. For Stage~2, total training rounds are $T = 500$ for HumanEval and $T = 1200$ for MATH, GSM8K, MBPP, and GAIA. During inference, we run five independent evaluations per test query; for each benchmark, we compute the overall metric (Solve Rate / pass@1) per run and report the average across runs. See Appendix~\ref{app:implementation} for complete hyperparameters and dataset statistics.

\subsection{Performance Analysis}
\label{subsec:main_results}
Table~\ref{tab:main_results} summarizes the performance of CB-Orchestrator compared to baseline methods on math reasoning and code generation benchmarks. The results demonstrate that \textbf{CB-Orchestrator consistently achieves the best performance across all benchmarks}, reaching a leading average score of 81.37\%. This surpasses the strongest autonomous baselines, MaAS and AFlow, by 1.58\% and 2.61\% respectively, and significantly outperforms other baselines.

\begin{table}[htbp]
\setlength{\tabcolsep}{6pt}
\centering
\renewcommand{\arraystretch}{1.07}
\setlength{\aboverulesep}{0pt} % 关键：去掉横线上的透明间距
\setlength{\belowrulesep}{0pt} % 关键：去掉横线下的透明间距
\caption{The performance of CB-Orchestrator compared to baseline methods on math reasoning and code generation benchmarks. Backbone is GPT-4o-mini for all experiments. CB-Orchestrator consistently achieves the best performance across all benchmarks.}
\label{tab:main_results}
\begin{tabular}{lccccc}
\toprule
\textbf{Method} & \textbf{MATH} & \textbf{GSM8K} & \textbf{MBPP} & \textbf{HumanEval} & \textbf{Avg}. \rule{0pt}{2.3ex}\\ \midrule
\textit{Basic Prompting} \rule{0pt}{2.5ex} \\
Vanilla & 46.29 & 87.45 & 71.83 & 87.08 & 73.16 \\
CoT \citep{CoT} & 46.40 & 87.10 & 71.83 & 88.13 & 73.37 \\
CoT SC (5-shot) \citep{Self-consistency} & 47.91 & 87.57 & 73.60 & 88.60 & 74.42 \\ \midrule
\textit{Fixed Multi-Agent Systems} \rule{0pt}{2.5ex} \\
LLM-Debate \citep{LLMDebate}  & 48.54 & 89.47 & 70.29 & 88.68 & 74.25 \\
LLM-Blender \citep{LLMBlender}  & 46.92 & 88.35 & 77.05 & 88.80 & 75.28 \\
DyLAN \citep{liu2024dynamic}  & 48.63 & 89.98 & 77.30 & 90.42 & 76.58 \\
AgentVerse \citep{AgentVerse}  & 47.35 & 89.91 & 74.28 & 89.29 & 75.21 \\ \midrule
\textit{Autonomous Multi-agent Systems} \rule{0pt}{2.5ex} \\
AutoAgents \citep{AutoAgents}  & 45.32 & 87.69 & 71.95 & 87.64 & 73.15 \\
GPTSwarm \citep{GPTSwarm} & 47.88 & 89.14 & 77.43 & 89.32 & 75.94 \\
ADAS \citep{ADAS} & 43.18 & 86.12 & 68.13 & 84.19 & 70.41 \\
AgentSquare \citep{AgentSquare} & 48.51 & 87.62 & 78.46 & 89.08 & 75.92 \\
AFlow \citep{AFlow} & 51.28 & 91.16 & 81.67 & 90.93 & 78.76 \\
MaAS \citep{MaAS} & 51.82 & 92.30 & 82.17 & 92.85 & 79.79 \\ 
\midrule
\rowcolor[gray]{0.93}
\textbf{CB-Orchestrator}  & \textbf{53.87} & \textbf{93.19} & \textbf{84.05} & \textbf{94.35} & \textbf{ 81.37} \rule{0pt}{2.3ex} \\ \bottomrule
\end{tabular}
\end{table}

On the GAIA benchmark (Table~\ref{tab:gaia_results} in Appendix~\ref{app:gaia_results}), \textbf{CB-Orchestrator achieves the highest average score (21.0\%)}, with particularly strong performance on L1 tasks (28.6\% solve rate).

\subsection{Cost Efficiency and Training Dynamics}
\label{subsec:efficiency}
\begin{figure}[htbp]
    \centering
    \includegraphics[width=0.8\linewidth]{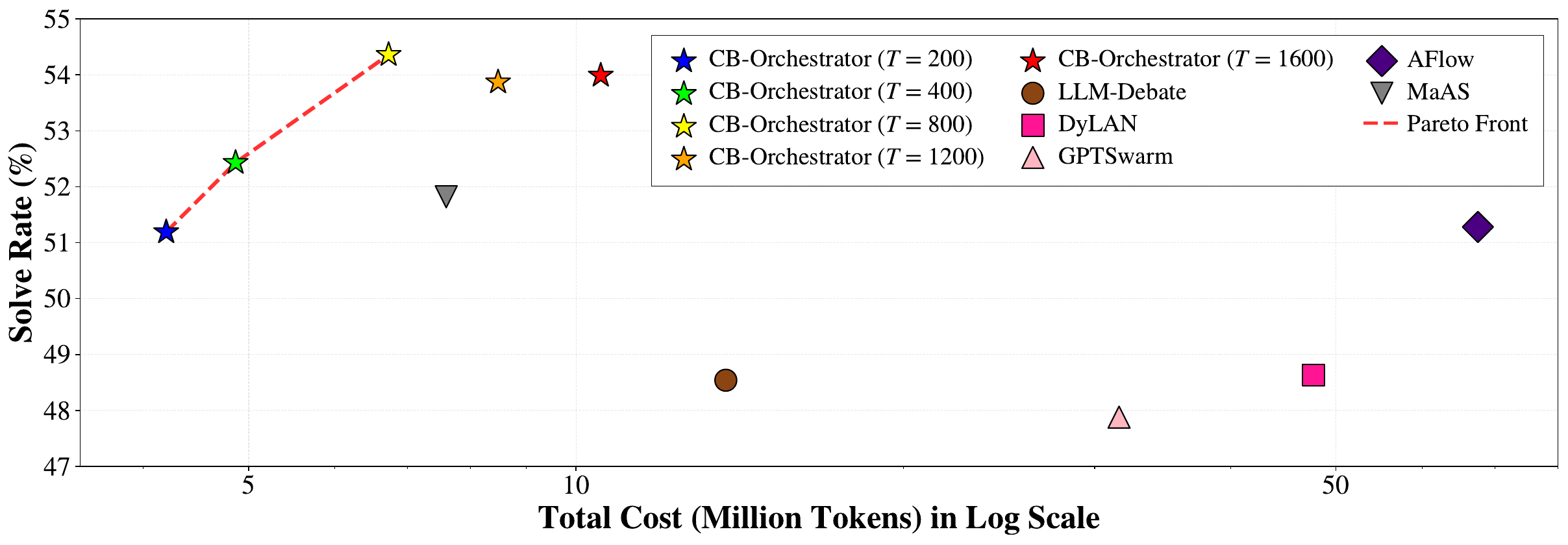} 
    \caption{Cost-Efficiency comparison on the MATH benchmark.}
    \label{fig:Cost_Solverate}
\end{figure}

A key advantage of CB-Orchestrator is its training-phase sample efficiency. To assess this, Figure~\ref{fig:Cost_Solverate} plots MATH solve rate against total token cost (training and test) for CB-Orchestrator at different $T$ and the baselines. The results place CB-Orchestrator on a superior Pareto front. At $T=400$, it reaches 52.43\% with about 4.28 million tokens, outperforming all baselines. At $T=800$, it raises the solve rate to 54.36\% while substantially reducing total token cost and keeping test-time overhead relatively low. Because the workflow pool is reusable, excluding its construction cost, $T=800$ \textbf{reduces training token cost by no less than 40.55\%} compared with baselines reporting training costs on MATH. \textbf{Including the one-time pool construction cost, the reduction remains at least 29.73\%} relative to the same baselines (Appendix~\ref{app:cost_breakdown}).

The training dynamics further show how these savings can be realized. Figure~\ref{fig:Cost_Solverate} indicates convergence before $T=800$, while Figure~\ref{fig:Itr_AverageReward} (Appendix~\ref{app:training_dynamics}) places it around round 650. In practice, the cumulative average reward provides a signal for when to stop training, allowing the reported cost reductions to be achieved.

\subsection{Ablation Study and Strategic Analysis}
\label{subsec:ablation}
\begin{figure}[h]
\centering
\includegraphics[width=0.7\linewidth]{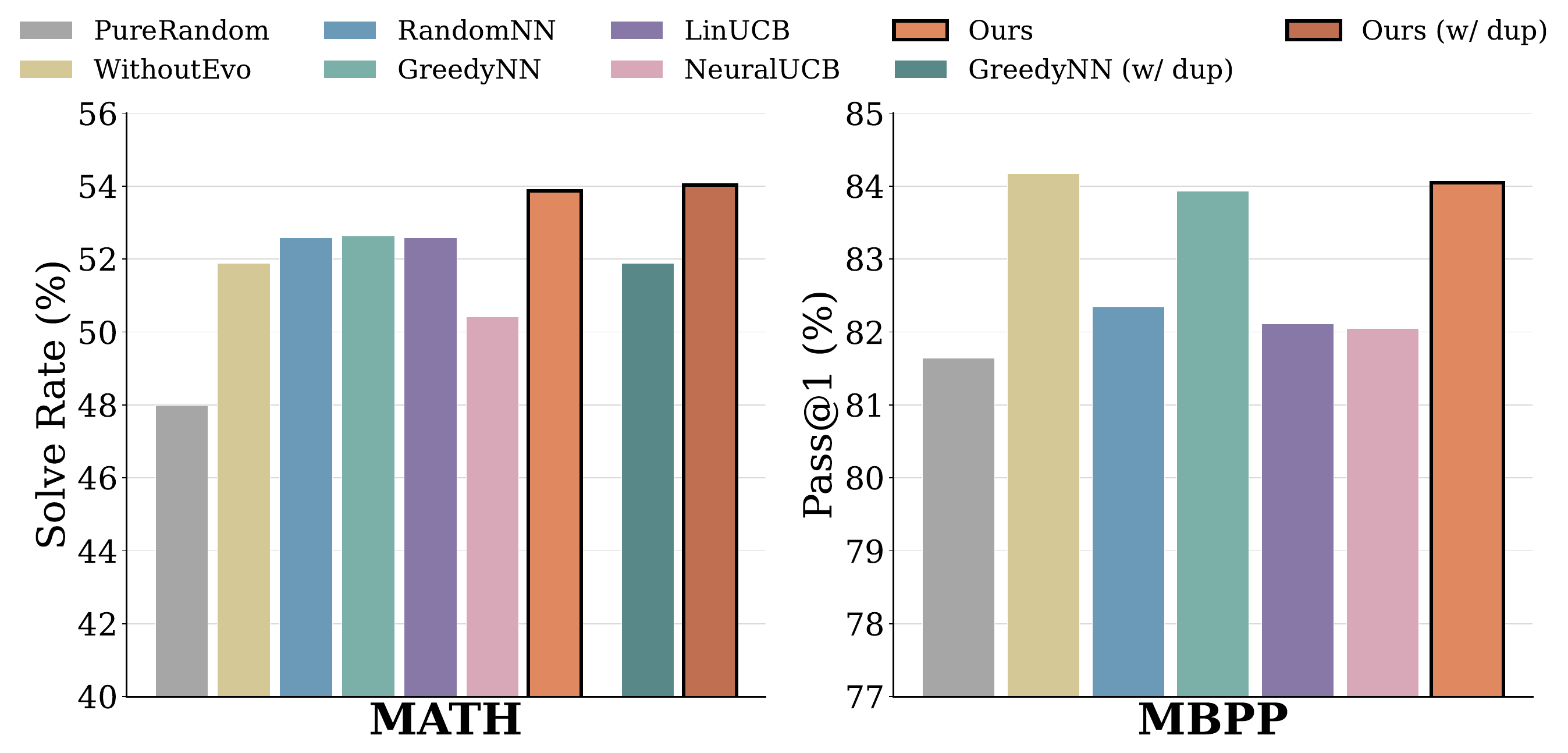}
\caption{Ablation results on MATH and MBPP. Black outlines indicate our proposed approaches.}
\label{fig:ablation_combined}
\end{figure}

To investigate the contribution of each component in CB-Orchestrator, we conduct an ablation study on the MATH and MBPP benchmarks. We first define the experimental variants to clarify the impact of each design choice: PureRandom selects workflows randomly from the pool to establish a baseline. RandomNN selects samples randomly for training the neural network. GreedyNN removes the exploration bonus to evaluate the necessity of the uncertainty based exploration term. LinUCB \citep{LinUCB} and NeuralUCB \citep{NeuralUCB} represent standard linear and neural bandit respectively. WithoutEvo removes the evolutionary search phase and performs selection over a static initial pool of workflows. The results are summarized in Figure~\ref{fig:ablation_combined}. Detailed numerical results are provided in Table~\ref{tab:ablation_results} (Appendix~\ref{app:ablation_results}).

As indicated in Figure~\ref{fig:ablation_combined}, \textbf{the workflow evolution mechanism provides a complementary benefit to the overall system}. Even the Without Evolution variant which performs bandit selection over a static initial workflow pool already achieves a strong average score of 68.03\%. By integrating the iterative evolution process, CB-Orchestrator further boosts the performance to 68.96\%. This result suggests that, while bandit-based selection serves as the primary driver of task adaptation, the evolution process acts as an effective optimization tool that refines workflow structures and further lifts the performance ceiling of the framework.

\textbf{Our HybridUCB exploration strategy performs better than all baseline selection strategies.} As shown in Figure~\ref{fig:ablation_combined}, all bandit-based variants significantly surpass the PureRandom baseline. This confirms that leveraging workflow and query features to workflow selection is essential for performance. Interestingly, GreedyNN achieves a high score of 68.28\%, which exceeds the performance of RandomNN, LinUCB, and NeuralUCB. This competitive performance of GreedyNN is due to our training protocol where re-exploring the same query-workflow pairs is disallowed. This constraint forces the model to maintain a degree of exploration capability even without an explicit bonus term.

The underperformance of LinUCB and NeuralUCB highlights the complexity of the adaptive selection task. LinUCB is 1.61\% lower than ours. It is hard for LinUCB to capture the intricate relationships between different dimensions of the embedding vectors. This limitation causes the model to settle on a single fixed workflow and prevents it from making adaptive selections based on the query context. NeuralUCB yields the lowest score among trained variants because its exploration assumption is too strong for our specific scenario. This leads to unreasonable exploration behavior that hinders stable learning. By contrast, our HybridUCB effectively balances exploitation via neural reward modeling with a reasonable exploration bonus to achieve best results.

\textbf{The exploration bonus becomes more critical when duplicate sampling is allowed.} We further conduct ablation experiments on MATH where the same (query, workflow) pairs can be evaluated multiple times during training. As shown in Figure~\ref{fig:ablation_combined}, when duplicate sampling is allowed (w/ dup), Ours achieves 54.03\% while GreedyNN drops to 51.89\%. This 2.14\% performance gap is larger than the 1.24\% gap observed in the no-duplicate setting, demonstrating that the exploration bonus in HybridUCB plays a more important role when the training protocol does not inherently enforce exploration through the no-duplicate constraint.

\subsection{Generalizability and Scalability Analysis}
\label{subsec:generalization}
We further examine CB-Orchestrator's generalizability across domains and LLM backbones.

\textbf{Cross-Domain Transfer.} We evaluate transfer between MATH and GSM8K in two settings. In \textbf{(i) Source Pool \& Source NN}, the evolved workflow pool and neural reward model trained on the source domain are applied to the target without refinement (zero-shot transfer). In \textbf{(ii) Source Pool \& Retrained NN}, we retain the source pool but train the reward model on the target using HybridUCB with $T=1200$. Table~\ref{tab:cross_domain_final} shows that even the zero-shot setting outperforms many baselines. Retraining raises MATH performance to 54.94\%, above the strongest baseline and our model trained on MATH. We hypothesize that GSM8K's larger training set may provide richer feedback for learning the selector.

\begin{table}[h]
\centering
\renewcommand{\arraystretch}{1.03}
\setlength{\aboverulesep}{0pt}
\setlength{\belowrulesep}{0pt}
\caption{Cross-domain transferability analysis. M and G denote MATH and GSM8K benchmarks, respectively. We evaluate the performance of CB-Orchestrator compared to baseline methods.}
\label{tab:cross_domain_final}
\begin{tabular}{lcc}
\toprule
\textbf{Method} & \textbf{M $\rightarrow$ G} & \textbf{G $\rightarrow$ M} \rule{0pt}{2.3ex} \\ \midrule
\multicolumn{3}{l}{\textit{Baseline Methods}} \rule{0pt}{2.5ex} \\
GPTSwarm & 89.96 & 45.18 \\
AFlow & 91.95 & 49.39 \\
MaAS & 92.80 & 51.02 \\ \midrule
\multicolumn{3}{l}{\textit{CB-Orchestrator (Ours)}} \rule{0pt}{2.5ex} \\
Source Pool \& Source NN & 89.72 & 53.62 \\
Source Pool \& Retrained NN & 92.75 & 54.94 \\ \bottomrule
\end{tabular}
\end{table}

\textbf{LLM Backbone Scalability.} To verify the scalability of CB-Orchestrator across different LLM backbones, we evaluate its performance using Llama-3.1-70b as the backbone. We conduct experiments on the MATH and HumanEval benchmarks, because the strongest baseline, MaAS \citep{MaAS}, reported results on these two datasets for this backbone. As summarized in Table~\ref{tab:backbone_robustness}, CB-Orchestrator achieves an average score of 64.24\%, outperforming the Vanilla baseline by a large margin and demonstrating an overall performance slightly better than MaAS. These results demonstrate that our CB-Orchestrator scales effectively to open-source LLMs.

\subsection{Case Study}
\label{subsec:Case Study}
Two MATH cases illustrate CB-Orchestrator's routing among generated workflows. On the training queries, workflow\_31 succeeds where workflow\_3 fails for a decimal-expansion problem, while the reverse holds for a trigonometric-curve problem. The selector assigns the corresponding test queries to the workflows that succeeded on the related training problems, and both answers are correct on the first attempt in all five test runs. Appendix~\ref{app:case_study} provides a detailed analysis.

\section{Conclusion}
\label{sec:conclusion}

We introduced CB-Orchestrator, a two-stage framework for adaptive workflow optimization for LLM agents. We decouple workflow construction from adaptive selection. Stage~1 builds a diverse workflow pool via execution-guided evolutionary search; Stage~2 casts selection as a contextual bandit and applies Hybrid Neural-Linear UCB for sample-efficient exploration over query-workflow pairs.

Experiments across five benchmarks demonstrate that CB-Orchestrator consistently outperforms strong baselines. On the MATH dataset, it reduces training-phase token overhead by at least 40.55\% and end-to-end training overhead by at least 29.73\%. Ablation studies confirm that both the evolutionary pool construction and the Hybrid UCB exploration contribute to performance gains. Furthermore, CB-Orchestrator exhibits strong cross-domain transferability and generalizes effectively to different LLM backbones. These results, together with the framework's natural ability to incorporate prior knowledge into the workflow pool, suggest that CB-Orchestrator offers a practical path toward deploying versatile, cost-efficient LLM agents across diverse real-world applications.

\subsection*{AI use statement}
We used generative AI tools to assist with drafting and editing the manuscript and with writing implementation code. The authors take responsibility for the final manuscript, code, and scientific claims, including AI-assisted content.

\subsection*{Reproducibility statement}
Section~\ref{sec:methodology} specifies the method, and Section~\ref{subsec:setup} describes the experimental setup and evaluation protocol. Appendix~\ref{app:implementation} provides dataset splits, hyperparameters, architecture and feature details, and the training workflow subset. Appendices~\ref{app:operator_library} and~\ref{app:prompts} specify the operators and workflow generation prompts.

\bibliography{example_paper}
\bibliographystyle{iclr2027_conference}

\appendix
\section{Implementation Details}
\label{app:implementation}

\subsection{Dataset Statistics}

Table~\ref{tab:datasets} summarizes the datasets used in our experiments. We use the validation split for training (Stage 1 workflow pool construction and Stage 2 bandit training) and the test split for evaluation (Stage 3 inference).

\begin{table}[ht]
\centering
\caption{Dataset statistics}
\label{tab:datasets}
% 用 tabularx 代替 tabular，宽度设为 \textwidth
\begin{tabularx}{0.99\textwidth}{lcccc >{\raggedright\arraybackslash}X} 
\toprule
\textbf{Dataset} & \textbf{Domain} & \textbf{Train} & \textbf{Test} & \textbf{Workflows} & \textbf{Operators} \\
\midrule
MATH & Math reasoning & 119 & 486 & 68 & Custom, ScEnsemble, Programmer \\
GSM8K & Math reasoning & 264 & 1055 & 72 & Custom, ScEnsemble, Programmer \\
HumanEval & Code generation & 33 & 131 & 31 & Custom, CustomCodeGenerate, ScEnsemble, Test \\
MBPP & Code generation & 86 & 341 & 81 & Custom, CustomCodeGenerate, ScEnsemble, Test \\
GAIA & Multi-tool QA & 93 & 301 & 36 & Custom, Reasoning, Decompose, Search, FileAnalysis, Programmer, ScEnsemble, FinalAnswer \\
\bottomrule
\end{tabularx}
\end{table}

\noindent\textbf{GSM8K Training Subset for Stage 1.} GSM8K has a relatively large training split. For workflow pool construction (Stage 1), we sample 100 training queries from the full GSM8K training split to generate candidate workflows and to compute pool coverage statistics.

\subsection{Hyperparameters}

Table~\ref{tab:hyperparameters} summarizes all hyperparameters used in CB-Orchestrator.

\begin{table}[ht]
\centering
\caption{Hyperparameters for CB-Orchestrator.}
\label{tab:hyperparameters}
\begin{tabular}{lcc|lcc}
\toprule
\textbf{Parameter} & \textbf{Sym.} & \textbf{Val.} & \textbf{Parameter} & \textbf{Sym.} & \textbf{Val.} \\
\midrule
\multicolumn{6}{c}{\textit{Stage 1: Workflow Pool Construction}} \\
\midrule
Evolution rounds & $N_{\text{evolve}}$ & 2 & Transfer rounds & $N_{\text{transfer}}$ & 2 \\
Gen. temperature & - & 1.0 & Exec. temperature & - & 0 \\
Similar queries & $k$ & 5\\
\midrule
\multicolumn{6}{c}{\textit{Stage 2: Bandit Training}} \\
\midrule
Embedding dim & $d$ & 2560 & Learning rate & $\eta$ & $10^{-5}$ \\
Hidden dim 1 & $h_1$ & 2048 & Weight decay & - & $10^{-3}$ \\
Hidden dim 2 & $h_2$ & 64 & Gradient clip & - & 1.0 \\
Exploration coef. & $\alpha$ & 0.1 & Epochs/batch & $E$ & 50 \\
Regularization & $\lambda$ & 0.5 & NN batch size & - & 16 \\
Total rounds & $T$ & See below & Explore rounds & $T_{\text{explore}}$ & 100 \\
Eval batch size & $B$ & 25 & Random seed & - & 42 \\
\midrule
\multicolumn{6}{c}{\textit{Stage 3: Inference}} \\
\midrule
Wilson confidence & $z$ & 3.891 & Min samples & $n_{\min}$ & 5 \\
Keep ratio & - & 0.5 & & & \\
\bottomrule
\end{tabular}
\end{table}

\noindent\textbf{Training Rounds per Dataset.} We use $T = 1200$ for MATH, GSM8K, MBPP, and GAIA, and $T = 500$ for HumanEval.

\subsection{Architecture and Feature Representation Details}
\label{app:architecture}

\textbf{Neural Network Architecture.}
The reward prediction network is a 3-layer MLP with pre-norm design and progressive dimensionality reduction:
\begin{align}
    h_0 &= \text{LN}(x) \in \mathbb{R}^{2d} \\
    h_1 &= \text{GELU}(\text{LN}(W_1 h_0 + b_1)) \in \mathbb{R}^{d_1} \\
    h_2 &= \text{GELU}(\text{LN}(W_2 h_1 + b_2)) \in \mathbb{R}^{d_2} \\
    f_\theta(x) &= \sigma(W_3 h_2 + b_3) \in [0, 1]
\end{align}
where $\text{LN}$ denotes layer normalization, $\text{GELU}$ is the activation function, and $\sigma$ is the sigmoid output. We use $d = 2560$ (embedding dimension), $d_1 = 2048$, and $d_2 = 64$, progressively compressing the representation from $2d = 5120$ to 1.

\textbf{Feature Representation.}
Workflow source code serves as a stable semantic summary of the problem-solving strategy. Each workflow contains: (1) the sequence of operators invoked (e.g., \texttt{Programmer}, \texttt{ScEnsemble}, \texttt{Test}), (2) their composition structure (sequential, branching, iterative), and (3) operator-specific parameters and prompts.

\textbf{Wilson Score Filtering.}
At inference time, we filter workflows using Wilson score confidence intervals to improve robustness. For each workflow $w$ with $n$ trials and $s$ successes during training, the Wilson lower bound is:
\begin{equation}
    \text{Wilson}(w) = \frac{s + \frac{z^2}{2}}{n + z^2} - \frac{z}{n + z^2} \sqrt{\frac{s(n-s)}{n} + \frac{z^2}{4}}
\end{equation}
where $z = 3.891$ corresponds to 99.99\% confidence. We require $n_{\min} = 5$ samples per workflow and keep only the top 50\% of workflows ranked by Wilson score, filtering out unreliable workflows with insufficient trials or consistently poor performance.

\subsection{Online Learning Mode}
\label{app:online_learning}

A natural concern is that the workflow pool $\mathcal{W}$, though diverse, is finite and constructed from training queries. When encountering queries that fall outside the pool's coverage---requiring reasoning patterns not captured by any existing workflow---the system can only select the ``least bad'' option. To address this limitation, CB-Orchestrator supports an optional \emph{online learning mode} that enables both model adaptation and dynamic pool expansion.

\textbf{Model Adaptation.}
After each inference, the model can continue learning from observed rewards:
\begin{equation}
    A \leftarrow A + x x^\top, \quad \mathcal{D} \leftarrow \mathcal{D} \cup \{(x, r)\}
\end{equation}
Periodic retraining of $f_\theta$ adapts the selection policy to distribution shifts in the query stream, improving performance on recurring query patterns.

\textbf{Coverage Gap Detection.}
For each query $q$, we compute a coverage score reflecting the pool's ability to handle it:
\begin{equation}
    \text{Coverage}(q) = \max_{w \in \mathcal{W}} f_\theta(x(q, w))
\end{equation}
A low coverage score (below threshold $\tau_{\text{gap}}$) combined with high uncertainty $u(x(q, w))$ across all workflows signals that $q$ likely requires a strategy absent from the current pool. We define the gap indicator:
\begin{equation}
    \text{Gap}(q) = \mathbb{I}\left[ \text{Coverage}(q) < \tau_{\text{gap}} \land \min_{w \in \mathcal{W}} u(x(q, w)) > u_{\text{thresh}} \right]
\end{equation}

\textbf{On-Demand Workflow Generation.}
When $\text{Gap}(q) = 1$, we invoke the same generation mechanism from Stage 1 using $q$ as the seed query: (1) generate an initial workflow tailored to $q$'s requirements, (2) evolve using execution feedback if the initial attempt fails, and (3) add successful workflows to the pool. The new workflow $w_{\text{new}}$ is integrated into the bandit:
\begin{equation}
    \mathcal{W} \leftarrow \mathcal{W} \cup \{w_{\text{new}}\}
\end{equation}
Since $w_{\text{new}}$ has no prior observations, its uncertainty $u(x(q, w_{\text{new}}))$ is naturally high, encouraging exploration during subsequent queries. This mechanism enables the system to continuously expand its capabilities rather than being constrained to the initial pool.

\textbf{Practical Considerations.}
On-demand generation incurs additional latency and cost, so it should be triggered judiciously. In practice, we recommend: (1) batching gap queries and generating workflows offline during low-traffic periods, (2) setting conservative thresholds ($\tau_{\text{gap}} \approx 0.3$, $u_{\text{thresh}}$ at the 90th percentile of training uncertainties) to avoid unnecessary generation, and (3) periodically pruning redundant workflows using the Wilson score filtering mechanism. This design ensures that the pool grows adaptively while maintaining computational efficiency.

\subsection{Workflow Pool Coverage and Training Subset}

\textbf{Pool Coverage.} Table~\ref{tab:pool_coverage} reports the coverage of our constructed workflow pools. In all benchmarks, the evaluator provides binary feedback $r \in \{0, 1\}$ (success/failure), so we use a fixed success threshold $\tau = 1$ without additional tuning. Here, ``Train Queries'' refers to the subset of training queries used in Stage 1 workflow pool construction for that dataset (e.g., we sample 100 training queries for GSM8K). Coverage is defined as the fraction of these Stage 1 training queries for which at least one workflow in the pool achieves success ($r \geq \tau$). High coverage indicates that the pool contains effective workflows for diverse query types.

\begin{table}[ht]
\centering
\caption{Workflow pool coverage across datasets.}
\label{tab:pool_coverage}
\begin{tabular}{lcccc}
\toprule
\textbf{Dataset} & \textbf{Train Queries} & \textbf{Covered} & \textbf{Coverage} & \textbf{Pool Size} \\
\midrule
MATH & 119 & 86 & 72.3\% & 68 \\
GSM8K & 100 & 98 & 98.0\% & 72 \\
HumanEval & 33 & 31 & 93.9\% & 31 \\
MBPP & 86 & 82 & 95.3\% & 81 \\
\bottomrule
\end{tabular}
\end{table}

The lower coverage on MATH (72.3\%) reflects the inherent difficulty of competition-level mathematics problems. Despite this, our bandit policy learns to select effective workflows for the covered queries.

\textbf{Training Workflow Subset.} For Stage 2 bandit training, we sample a subset of $K \approx 30$ workflows from the full pool rather than using all workflows. This design choice is motivated by several observations:
\begin{itemize}
    \item \textbf{Empirical sufficiency}: Our experiments show that 30 workflows provide sufficient diversity to achieve strong performance, as many workflows in the pool have similar strategies and near-identical success patterns.
    \item \textbf{Training efficiency}: With $K$ arms and $T$ training rounds, the mean number of observations per arm is $T/K$, although the adaptive policy may allocate observations unevenly. A smaller candidate set can provide more observations per arm under a fixed training budget.
    \item \textbf{Inference filtering}: The Wilson score filtering in Stage 3 (Section~\ref{sec:inference}) further reduces the active workflow set, so training on a subset does not limit final performance.
\end{itemize}
In practice, the number of training workflows can be adjusted based on the training budget and pool diversity. For datasets with smaller pools (e.g., HumanEval with 31 workflows), we use all available workflows.

\subsection{Cost Breakdown and Accounting}
\label{app:cost_breakdown}

To ensure fair comparison, we provide a detailed cost breakdown for all methods on the MATH benchmark. Table~\ref{tab:cost_breakdown} reports token consumption across different phases.

\begin{table}[ht]
\centering
\caption{Detailed cost breakdown on MATH benchmark (all values in millions of tokens).}
\label{tab:cost_breakdown}
\small
\begin{tabular}{lccccc}
\toprule
\textbf{Method} & \textbf{Pool Gen.} & \textbf{Training} & \textbf{Inference} & \textbf{Total} & \textbf{Solve \%} \\
\midrule
LLM-Debate & -- & -- & 13.73 & 13.73 & 48.54 \\
DyLAN & -- & 38.30 & 9.39 & 47.68 & 48.63 \\
GPTSwarm & -- & 27.70 & 3.89 & 31.59 & 47.88 \\
AFlow & -- & 62.88 & 4.66 & 67.54 & 51.28 \\
MaAS & -- & 5.43 & 2.16 & 7.60 & 51.82 \\
\midrule
CB-Orch ($T$=200) & 0.59 & 0.78 & 2.24 & 3.61 & 51.19 \\
CB-Orch ($T$=400) & 0.59 & 1.53 & 2.16 & 4.28 & 52.43 \\
CB-Orch ($T$=800) & 0.59 & 3.23 & 2.32 & 6.14 & 54.36 \\
CB-Orch ($T$=1200) & 0.59 & 5.00 & 2.30 & 7.89 & 53.87 \\
CB-Orch ($T$=1600) & 0.59 & 6.91 & 2.45 & 9.95 & 53.99 \\
\bottomrule
\end{tabular}
\end{table}

\textbf{Cost Accounting Principles.} We apply consistent accounting across all methods:
\begin{itemize}
    \item \textbf{All LLM calls counted}: Training, inference, workflow generation, and any internal evaluator calls are included.
    \item \textbf{Failed executions}: Executions that fail or raise exceptions still incur token costs; we count all tokens consumed before failure.
    \item \textbf{Input and output tokens}: Both prompt tokens and completion tokens are summed. Table~\ref{tab:cost_io_breakdown} provides the input/output breakdown.
\end{itemize}

\begin{table}[ht]
\centering
\caption{Input/output token breakdown on MATH (millions of tokens).}
\label{tab:cost_io_breakdown}
\small
\begin{tabular}{lcccc}
\toprule
\textbf{Method} & \textbf{Train Input} & \textbf{Train Output} & \textbf{Infer Input} & \textbf{Infer Output} \\
\midrule
LLM-Debate & -- & -- & 3.28 & 10.46 \\
DyLAN & 22.15 & 16.15 & 6.08 & 3.30 \\
GPTSwarm & 21.33 & 6.37 & 3.11 & 0.79 \\
AFlow & 33.83 & 29.05 & 2.51 & 2.15 \\
MaAS & 3.05 & 2.38 & 1.31 & 0.85 \\
\midrule
CB-Orch ($T$=1200) & 3.24 & 1.76 & 1.60 & 0.70 \\
\bottomrule
\end{tabular}
\end{table}

\textbf{Workflow Pool Reusability.} The workflow pool generation cost (0.59M tokens for MATH) is a one-time expense that can be amortized across multiple deployments. If we exclude pool generation, CB-Orchestrator ($T$=800) achieves 54.36\% solve rate with only 5.55M tokens, representing a 26.97\% total token cost (including training and test phase) reduction compared to MaAS.

\subsection{Attribution}
\label{app:attribution}

Parts of the code used in our experiments are adapted from the open-source implementation of AFlow~\citep{AFlow} and MaAS~\citep{MaAS}

\subsection{Training Dynamics}
\label{app:training_dynamics}

\begin{figure}[ht]
    \centering
    \includegraphics[width=0.45\linewidth]{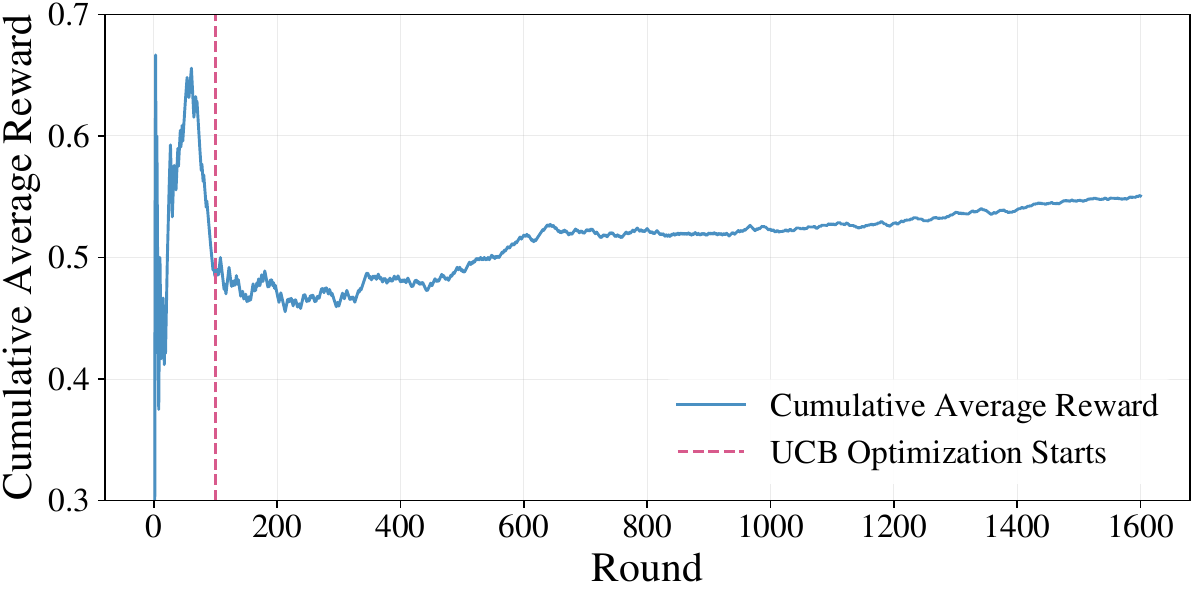}
    \caption{Cumulative average reward during bandit training on MATH. The curve shows rapid convergence at around Round 650, indicating that the bandit policy has learned effective workflow selection.}
    \label{fig:Itr_AverageReward}
\end{figure}

\subsection{Operator Library}
\label{app:operator_library}

Following AFlow~\citep{AFlow}, we define operators as reusable atomic building blocks that encapsulate common LLM interaction patterns. Each operator has a standardized interface with typed inputs and outputs, enabling modular composition. We design domain-specific operator libraries for different task types. Table~\ref{tab:operators_math} shows operators for math reasoning (MATH, GSM8K), Table~\ref{tab:operators_code} for code generation (HumanEval, MBPP), and Table~\ref{tab:operators_qa} for multi-tool QA (GAIA).

\begin{table}[ht]
\centering
\caption{Operators for math reasoning benchmarks (MATH, GSM8K).}
\label{tab:operators_math}
\small
\begin{tabular}{lp{10cm}}
\toprule
\textbf{Operator} & \textbf{Description \& Interface} \\
\midrule
\texttt{Custom} & General-purpose LLM call with customizable prompts. Flexible for planning, reflection, decomposition, formatting, etc. \\
& \texttt{(input: str, instruction: str) -> \{response: str\}} \\
\midrule
\texttt{ScEnsemble} & Self-consistency voting across multiple solutions. Requires 3+ solutions for effective voting. \\
& \texttt{(solutions: List[str], problem: str) -> \{response: str\}} \\
\midrule
\texttt{Programmer} & Generates and executes Python code in a sandbox. Auto-retries on errors. Returns computed answer. \\
& \texttt{(problem: str, analysis: str) -> \{code: str, output: str\}} \\
\bottomrule
\end{tabular}
\end{table}

\begin{table}[ht]
\centering
\caption{Operators for code generation benchmarks (HumanEval, MBPP).}
\label{tab:operators_code}
\small
\begin{tabularx}{\linewidth}{l>{\raggedright\arraybackslash}X}
\toprule
\textbf{Operator} & \textbf{Description \& Interface} \\
\midrule
\texttt{Custom} & General-purpose LLM call for reasoning about code structure and logic. \\
& \texttt{(input: str, instruction: str) -> \{response: str\}} \\
\midrule
\texttt{CustomCodeGenerate} & Specialized code generation with entry point awareness. \\
& \texttt{(problem: str, entry\_point: str, instruction: str) -> \{response: str\}} \\
\midrule
\texttt{ScEnsemble} & Self-consistency voting for code solutions. \\
& \texttt{(solutions: List[str], problem: str) -> \{response: str\}} \\
\midrule
\texttt{Test} & Executes test cases and iteratively refines code. Retries up to 3 times on failures. \\
& \texttt{(problem: str, solution: str, entry\_point: str) -> \{result: bool, solution: str\}} \\
\bottomrule
\end{tabularx}
\end{table}

\begin{table}[ht]
\centering
\caption{Operators for multi-tool QA benchmark (GAIA).}
\label{tab:operators_qa}
\small
\begin{tabular}{lp{10cm}}
\toprule
\textbf{Operator} & \textbf{Description \& Interface} \\
\midrule
\texttt{Custom} & General-purpose LLM call for flexible reasoning and formatting. \\
& \texttt{(input: str, instruction: str) -> \{response: str\}} \\
\midrule
\texttt{Reasoning} & Multi-step structured reasoning. Can receive context from other operators. \\
& \texttt{(question: str, context: str) -> \{thought: str, answer: str\}} \\
\midrule
\texttt{Decompose} & Breaks complex questions into sub-tasks with dependency tracking. \\
& \texttt{(question: str, context: str) -> \{analysis: str, subtasks: List, final\_aggregation: str\}} \\
\midrule
\texttt{Search} & Iterative web search with query generation, page fetching, and answer extraction. \\
& \texttt{(question: str) -> \{answer: str\}} \\
\midrule
\texttt{FileAnalysis} & Extracts information from attachments (PDF, Excel, CSV, images, text, JSON). \\
& \texttt{(question: str, file\_path: str) -> \{file\_info: str\}} \\
\midrule
\texttt{Programmer} & Python code execution for computational tasks. \\
& \texttt{(question: str, context: str) -> \{code: str, output: str\}} \\
\midrule
\texttt{ScEnsemble} & Self-consistency voting across multiple solutions. \\
& \texttt{(solutions: List[str], problem: str) -> \{response: str\}} \\
\midrule
\texttt{FinalAnswer} & Mandatory final step for answer formatting per QA evaluation rules. \\
& \texttt{(question: str, analysis: str) -> \{thought: str, final\_answer: str\}} \\
\bottomrule
\end{tabular}
\end{table}

\clearpage
\subsection{Workflow Example}

Below is an example workflow generated for MATH problems. The workflow uses a three-step structure: decomposition, code-based solution, and LaTeX formatting.

\begin{tcolorbox}[colback=teal!3, colframe=teal!60, title=Example Workflow for MATH (Python), fonttitle=\bfseries\small]
\small
\begin{verbatim}
class Workflow:
    def __init__(self, name, llm_config, dataset):
        self.name = name
        self.llm = create_llm_instance(llm_config)
        self.custom = op.Custom(self.llm)
        self.programmer = op.Programmer(self.llm)

    async def __call__(self, problem: str):
        # Step 1: Decompose the problem
        decomposition = await self.custom(
            input=problem,
            instruction="Break down the problem into steps...")

        # Step 2: Solve using code execution
        solution = await self.programmer(
            problem=decomposition['response'],
            analysis="Write a solve() function...")

        # Step 3: Format answer in LaTeX
        formatted = await self.custom(
            input=f"Solution: {solution['output']}",
            instruction="Format as \\boxed{answer}...")

        return formatted['response'], self.llm.get_cost()
\end{verbatim}
\end{tcolorbox}

\noindent This workflow demonstrates the typical operator composition pattern: \texttt{Custom} handles flexible text processing (decomposition, formatting), while \texttt{Programmer} executes code for numerical computation.

\subsection{Workflow Generation Prompts}
\label{app:prompts}

We provide the complete prompts used in Stage 1 for workflow generation and evolution. Variables in \texttt{\{braces\}} are filled at runtime.

\subsubsection{Initial Generation System Prompt}

\begin{tcolorbox}[breakable, title after break={WORKFLOW\_GENERATE\_SYSTEM (continued)}, colback=teal!3, colframe=teal!60, fonttitle=\bfseries\small, title=WORKFLOW\_GENERATE\_SYSTEM]
\small\ttfamily
You are building a Graph to solve \{type\} problems.\\
Please construct a complete and correct workflow graph.\\[0.5em]
**IMPORTANT: When initializing operators in \_\_init\_\_, you must pass self.llm as the first argument. Check the 'initialization' field in operator\_description for the correct syntax.**\\[0.5em]
**CRITICAL - Operator Call Syntax:**\\
- Operators are callable objects (they implement \_\_call\_\_). Call them DIRECTLY on the instance:\\
\hspace*{1em}- CORRECT: `await self.search(question=problem)`, \\`await self.file\_analysis(question=problem, file\_path=file\_path)`\\
\hspace*{1em}- WRONG: `await self.search.search(...)`,\\ `await self.file\_analysis.file\_analysis(...)`\\
- This applies to ALL operators: custom, reasoning, decompose, search, file\_analysis, programmer, sc\_ensemble, final\_answer\\[0.5em]
**CRITICAL - Async Result Handling:**\\
- You CANNOT access dict keys directly after await in one expression!\\
- WRONG: `return await self.final\_answer(...)['final\_answer']` (causes "coroutine was never awaited" error)\\
- CORRECT: First store the result, then access keys:\\
\hspace*{1em}result = await self.final\_answer(...)\\
\hspace*{1em}return result['final\_answer'], ...\\[0.5em]
Here's an example of using the `custom` method in graph:\\
\hspace*{1em}CUSTOM\_PROMPT = "Your instruction here"\\
\hspace*{1em}response = await self.custom(input=problem, instruction=CUSTOM\_PROMPT)\\
In custom, the input and instruction are directly concatenated (instruction+input).\\
You may construct `input` by concatenating problem and strings previously generated by other operators, such as problem statement, previous reasoning, candidate solutions, test feedback, etc. using f-strings.\\
The Custom operator is very flexible and can be used multiple times with different prompts. The function of Custom method can be Planning, Reflection, Decomposition, Synthesis and whatever you want.\\
**IMPORTANT - Do not use Programmer unless necessary.**\\[0.5em]
**IMPORTANT: ScEnsemble requires MULTIPLE DIFFERENT solutions (typically 3+) to the SAME problem for voting. Do NOT use ScEnsemble with only one solution.**\\[0.5em]
**IMPORTANT - Answer Format Standardization:**\\
- **For math problems**: MUST add a final Custom node to format the answer using proper LaTeX format:\\
\hspace*{1em}FORMAT\_PROMPT = '''Given the problem and solution, extract only the final answer and format it as \textbackslash boxed\{\{answer\}\}.\\
\hspace*{1em}IMPORTANT formatting rules:\\
\hspace*{1em}- Use LaTeX \textbackslash frac\{\{a\}\}\{\{b\}\} for fractions, NOT a/b\\
\hspace*{1em}- Use LaTeX \textbackslash sqrt\{\{x\}\} for square roots, NOT sqrt(x)\\
\hspace*{1em}- Use \textbackslash pi for pi, NOT decimal approximations'''\\
- **For code problems**: Test operator already returns properly formatted code, so DO NOT add formatting after Test.\\[0.5em]
**Under no circumstances should Graph output None for any field.**
\end{tcolorbox}

\subsubsection{Initial Generation User Prompt}

\begin{tcolorbox}[breakable, title after break={WORKFLOW\_GENERATE\_INPUT (continued)}, colback=teal!3, colframe=teal!60, fonttitle=\bfseries\small, title=WORKFLOW\_GENERATE\_INPUT]
\small\ttfamily
Here is a problem example from the \{type\} domain:\\
<problem>\\
\{problem\}\\
</problem>\\[0.5em]
Available operators:\\
\{operator\_description\}\\[0.5em]
Please generate a complete workflow graph that can solve \{type\} problems.\\
**CRITICAL: Design the workflow to be generalizable across the \{type\} problem category, NOT just tailored to this specific instance.**\\[0.5em]
The graph must follow this exact template structure:\\[0.3em]
from typing import Literal\\
import asyncio\\
import workspace\_bandit.\{dataset\}.workflows.template.operator as op\\
from scripts.async\_llm import create\_llm\_instance\\
from scripts.evaluator import DatasetType\\[0.3em]
class Workflow:\\
\hspace*{1em}def \_\_init\_\_(self, name: str, llm\_config, dataset: DatasetType) -> None:\\
\hspace*{2em}self.name = name\\
\hspace*{2em}self.dataset = dataset\\
\hspace*{2em}self.llm = create\_llm\_instance(llm\_config)\\
\hspace*{2em}\# All operators pre-initialized:\\
\hspace*{2em}\{operator\_init\}\\[0.3em]
\hspace*{1em}async def \_\_call\_\_(self, problem: str\{entry\_point\_param\}):\\
\hspace*{2em}\# Your workflow logic here\\
\hspace*{2em}return answer, self.llm.get\_usage\_summary()["total\_cost"]\\[0.5em]
Output ONLY the complete Python code in <graph></graph> tags. DO NOT use markdown code blocks.
\end{tcolorbox}

\subsubsection{Evolution System Prompt}

\begin{tcolorbox}[breakable, title after break={WORKFLOW\_EVOLVE\_SYSTEM (continued)}, colback=teal!3, colframe=teal!60, fonttitle=\bfseries\small, title=WORKFLOW\_EVOLVE\_SYSTEM]
\small\ttfamily
You are an expert at debugging and improving workflow graphs for \{type\} problems.\\[0.5em]
A workflow failed to solve a problem. Your task is to analyze the failure and create an IMPROVED workflow.\\[0.5em]
\#\#\#\#\#\#\#\#\#\#\#\#\#\#\#\#\#\#\#\#\#\#\#\#\#\#\#\#\#\#\#\#\#\#\#\#\\
NOTICE:\\
**VERY CRITICAL ANTI-OVERFITTING RULES:**\\
1. You MUST NOT copy or embed any problem-specific tokens, values, strings, or names into your prompts/logic/code\\
2. You MUST NOT hardcode constants that come from this specific problem (e.g., specific numbers, names, or strings from the problem)\\
3. Your fix must work for ALL similar problems, not just this one\\
4. Focus on improving the GENERAL strategy, not solving this specific instance\\
\#\#\#\#\#\#\#\#\#\#\#\#\#\#\#\#\#\#\#\#\#\#\#\#\#\#\#\#\#\#\#\#\#\#\#\#\\[0.5em]
**IMPORTANT: When initializing operators in \_\_init\_\_, you must pass self.llm as the first argument.**\\[0.5em]
**Improvement Approach:**\\
Analyze the execution trace and error information to identify the root cause.\\
Based on your analysis, decide what changes are needed - this could include:\\
- **Modifying prompts**\\
- Adjusting workflow structure\\
- Adding/removing operators\\[0.5em]
**Under no circumstances should Graph output None for any field.**
\end{tcolorbox}

\subsubsection{Evolution User Prompt (Math/General)}

\begin{tcolorbox}[breakable, title after break={WORKFLOW\_EVOLVE\_INPUT\_MATH (continued)}, colback=teal!3, colframe=teal!60, fonttitle=\bfseries\small, title=WORKFLOW\_EVOLVE\_INPUT\_MATH]
\small\ttfamily
\#\# Previous Workflow (Failed):\\
```python\\
\{previous\_workflow\}\\
```\\[0.3em]
\#\# Problem That Failed:\\
<problem>\\
\{problem\}\\
</problem>\\[0.3em]
\#\# Execution Trace (Operator Inputs/Outputs):\\
\{execution\_trace\}\\[0.3em]
\#\# Error Information:\\
\{error\_info\}\\[0.3em]
\#\# Expected Answer:\\
\{expected\_output\}\\[0.3em]
\#\# Workflow's Predicted Answer:\\
\{actual\_output\}\\[0.3em]
\#\# Available Operators:\\
\{operator\_description\}
\end{tcolorbox}

\subsubsection{Evolution User Prompt (Code)}

\begin{tcolorbox}[breakable, title after break={WORKFLOW\_EVOLVE\_INPUT\_CODE (continued)}, colback=teal!3, colframe=teal!60, fonttitle=\bfseries\small, title=WORKFLOW\_EVOLVE\_INPUT\_CODE]
\small\ttfamily
\#\# Previous Workflow (Failed):\\
```python\\
\{previous\_workflow\}\\
```\\[0.3em]
\#\# Problem That Failed:\\
<problem>\\
\{problem\}\\
</problem>\\[0.3em]
\#\# Execution Trace (Operator Inputs/Outputs):\\
\{execution\_trace\}\\[0.3em]
\#\# Error Information:\\
\{error\_info\}\\[0.3em]
\#\# Correct Code (Reference Solution):\\
```python\\
\{correct\_code\}\\
```\\[0.3em]
\#\# Wrong Code (Workflow Generated):\\
```python\\
\{wrong\_code\}\\
```\\[0.3em]
\#\# Test Case Results:\\
\{test\_case\_results\}\\[0.3em]
\#\# Available Operators:\\
\{operator\_description\}
\end{tcolorbox}

\subsubsection{Evolution User Prompt (QA)}

\begin{tcolorbox}[breakable, title after break={WORKFLOW\_EVOLVE\_INPUT\_QA (continued)}, colback=teal!3, colframe=teal!60, fonttitle=\bfseries\small, title=WORKFLOW\_EVOLVE\_INPUT\_QA]
\small\ttfamily
\#\# Previous Workflow (Failed):\\
```python\\
\{previous\_workflow\}\\
```\\[0.3em]
\#\# Problem That Failed:\\
<problem>\\
\{problem\}\\
</problem>\\[0.3em]
\#\# Execution Trace (Operator Inputs/Outputs):\\
\{execution\_trace\}\\[0.3em]
\#\# Error Information:\\
\{error\_info\}\\[0.3em]
\#\# Expected Answer:\\
\{expected\_output\}\\[0.3em]
\#\# Workflow's Predicted Answer:\\
\{actual\_output\}\\[0.3em]
\#\# Available Operators:\\
\{operator\_description\}\\[0.5em]
\#\# QA-Specific Guidance:\\
- If file\_path is provided, call FileAnalysis FIRST - it returns 'file\_info' (extracted info, NOT an answer)\\
- FileAnalysis -> Reasoning (with file\_info as context) -> FinalAnswer is the standard file-based pipeline\\
- For web-based questions, use Search (check 'answer' field - empty string means search failed)\\
- For computational questions on file data, use FileAnalysis -> Programmer (pass file\_info as context)\\
- For factual questions, prefer Search over Reasoning when possible\\
- Always use FinalAnswer as the LAST step to format the answer correctly
\end{tcolorbox}

\subsubsection{Evolution Task Prompt}

\begin{tcolorbox}[breakable, title after break={WORKFLOW\_EVOLVE\_TASK (continued)}, colback=teal!3, colframe=teal!60, fonttitle=\bfseries\small, title=WORKFLOW\_EVOLVE\_TASK]
\small\ttfamily
---\\[0.3em]
**Your Task:**\\
1. Analyze WHY the workflow failed\\
2. Design an IMPROVED workflow that fixes the issue\\
3. **CRITICAL**: Ensure your fix is GENERALIZABLE - it must work for similar problems, not just this one\\[0.5em]
**CRITICAL - Operator Call Syntax:**\\
- Operators are callable objects. Call them directly: `await self.search(...)`, NOT `await self.search.search(...)`\\
- WRONG: `await self.file\_analysis.file\_analysis(...)`, `await self.reasoning.reasoning(...)`\\
- CORRECT: `await self.file\_analysis(...)`, `await self.reasoning(...)`\\[0.5em]
The graph must follow this exact template structure:\\[0.3em]
from typing import Literal\\
import asyncio\\
import workspace\_bandit.\{dataset\}.workflows.template.operator as op\\
from scripts.async\_llm import create\_llm\_instance\\
from scripts.evaluator import DatasetType\\[0.3em]
class Workflow:\\
\hspace*{1em}def \_\_init\_\_(self, name: str, llm\_config, dataset: DatasetType) -> None:\\
\hspace*{2em}self.name = name\\
\hspace*{2em}self.dataset = dataset\\
\hspace*{2em}self.llm = create\_llm\_instance(llm\_config)\\
\hspace*{2em}\# Initialize operators using op.OperatorName(self.llm)\\[0.3em]
\hspace*{1em}async def \_\_call\_\_(self, problem: str\{entry\_point\_param\}):\\
\hspace*{2em}\# Your improved workflow implementation\\
\hspace*{2em}return answer, self.llm.get\_usage\_summary()["total\_cost"]\\[0.5em]
Output a brief analysis in <analysis></analysis> tags (root cause + fix strategy), then the complete Python code in <graph></graph> tags.
\end{tcolorbox}

\section{Theoretical Discussion}
\label{app:theory}

\subsection{Design Rationale and Theoretical Positioning}

We begin with an honest assessment of our Hybrid Neural-Linear UCB design. Our approach is a \textbf{practical design choice} that prioritizes stability and computational efficiency over theoretical optimality. Unlike NeuralUCB~\citep{NeuralUCB} or Neural-LinUCB, which compute uncertainty in the neural network's learned representation space and enjoy formal regret guarantees under specific assumptions (e.g., the neural tangent kernel regime), our method decouples reward prediction from uncertainty quantification by using the \emph{original embedding space} for exploration bonuses.

This decoupling means we cannot directly inherit the $\tilde{O}(\sqrt{T})$ regret bounds from neural bandit theory. However, we argue that this design is well-motivated for our application, and we provide intuitive justification grounded in our smoothness assumption.

\subsection{Intuitive Justification via Lipschitz Smoothness}

The key intuition is that \textbf{under Assumption~\ref{assumption:smoothness} (reward smoothness), feature-space novelty can guide exploration toward less-covered query--workflow pairs}.

\textbf{Lipschitz Continuity Relates Nearby Expected Rewards.}
Assumption~\ref{assumption:smoothness} states that the expected reward function $f^*(x) = \mathbb{E}[r|x]$ is $L$-Lipschitz continuous:
\begin{equation}
    |f^*(x_1) - f^*(x_2)| \leq L \cdot \|x_1 - x_2\|
\end{equation}
Thus, nearby contexts have similar expected rewards. This bound does not directly constrain the error of a learned predictor or account for noise in an observed reward.

\textbf{Linear UCB Uncertainty Captures Feature-Space Novelty.}
The linear UCB uncertainty $u(x) = \sqrt{x^\top A^{-1} x}$ measures how ``novel'' the feature vector $x$ is relative to previously observed features. Geometrically, $u(x)$ is large when $x$ lies in directions that have been under-explored, and small when $x$ aligns with well-covered directions in the feature space.

\textbf{Connecting Feature-Space Novelty to Exploration.}
The quantity $u(x)$ reflects coverage in the original feature space: it is larger in directions less represented by observed contexts. Together with reward smoothness, this motivates exploring less-covered query--workflow pairs to collect feedback that may help the predictor generalize. However, $u(x)$ is not a calibrated confidence interval for $f_\theta(x)$ and does not by itself upper-bound $|f_\theta(x)-f^*(x)|$; we use it as a practical exploration signal.

\textbf{Informal Exploration Intuition.}
Let $\epsilon_t = |f_\theta(x_t) - f^*(x_t)|$ denote the neural network's prediction error at round $t$. As observations accumulate in less-covered regions, the reward predictor gains data that may reduce this error. The exploration bonus $\alpha \cdot u(x)$ encourages visiting such regions, which:
\begin{enumerate}
    \item Provides training data that may improve reward prediction in those regions;
    \item Increases the chance of identifying high-reward arms in less-explored areas.
\end{enumerate}
This intuition is consistent with the empirical performance gains observed in our experiments.

\subsection{Comparison with Neural Bandits}

Our approach differs fundamentally from NeuralUCB~\citep{NeuralUCB} and Neural-LinUCB in how uncertainty is quantified.

\textbf{NeuralUCB} computes uncertainty using the gradient of the neural network's output:
\begin{equation}
    u_{\text{NeuralUCB}}(x) = \sqrt{g(x)^\top Z^{-1} g(x)}, \quad g(x) = \nabla_{\theta_L} f_\theta(x)
\end{equation}
where $Z = \lambda I + \sum_t g(x_t) g(x_t)^\top$. This approach has formal regret guarantees under the neural tangent kernel (NTK) regime, but requires: (1) computing and storing gradients for each sample, (2) maintaining a $p \times p$ matrix where $p$ is the last-layer dimension, and (3) assuming the network operates in the NTK regime (wide networks, small learning rates).

\textbf{Our Hybrid Approach} uses the original feature space for uncertainty:
\begin{equation}
    u_{\text{Hybrid}}(x) = \sqrt{x^\top A^{-1} x}, \quad A = \lambda I + \sum_t x_t x_t^\top
\end{equation}

\textbf{Trade-offs.} Our design sacrifices theoretical guarantees for practical benefits:
\begin{itemize}
    \item \textbf{Computational efficiency}: No gradient computation or storage; the precision matrix $A$ is updated with simple outer products.
    \item \textbf{Training flexibility}: The neural network can be trained with any optimizer, learning rate, or architecture without affecting uncertainty estimates.
    \item \textbf{Stability}: Linear UCB uncertainty is well-behaved regardless of neural network training dynamics, avoiding gradient explosion or vanishing issues.
    \item \textbf{Interpretability}: The uncertainty directly reflects feature-space coverage, making exploration behavior predictable and tunable via the single parameter $\alpha$.
\end{itemize}

The cost is that our uncertainty estimate may be less accurate when the reward function is highly non-linear in the original feature space. However, our empirical results (Table~\ref{tab:ablation_results}) show that NeuralUCB underperforms our hybrid approach on this task, suggesting that the stability benefits outweigh the theoretical advantages of gradient-based uncertainty for workflow orchestration.

\subsection{Sample-Efficiency Intuition}

For a training set of $n$ queries and a pool of $K$ workflows, exhaustively evaluating every query--workflow pair would require $nK$ evaluations. CB-Orchestrator instead observes rewards for selected pairs and fits a shared reward model. Here, $nK$ counts the possible pairs; it is not a sample-complexity bound.

The joint embedding $x(q,w) = [\phi(q); \phi(w)]$ lets the model share information across pairs. Under the smoothness assumption (Assumption~\ref{assumption:smoothness}), nearby embeddings can have similar expected rewards. The precision matrix $A$ tracks feature-space coverage for the linear exploration bonus: observing one feature vector lowers this bonus in directions it covers. This describes the exploration mechanism, but does not establish a finite-sample guarantee for the neural reward model or the resulting policy. We therefore assess evaluation cost and performance empirically (Figure~\ref{fig:Cost_Solverate}).

\subsection{Computational Complexity and Scalability}

We analyze the computational costs of our Hybrid Neural-Linear UCB and discuss scalability to larger settings.

\textbf{Current Setting.} With embedding dimension $2d = 5120$, the precision matrix $A \in \mathbb{R}^{5120 \times 5120}$ requires $5120^2 \times 4 \approx 100$MB (float32). We avoid explicit matrix inversion by using Cholesky decomposition: decompose $A = LL^\top$, then solve $Ly = x$ for $y$, giving $u(x) = \|y\|_2$. This reduces per-query complexity from $O(d^3)$ (matrix inversion) to $O(d^2)$ (triangular solve). The Cholesky factor $L$ is recomputed once per batch after updating $A$, amortizing the $O(d^3)$ decomposition cost across $B$ queries. Table~\ref{tab:complexity} summarizes the per-operation costs.

\begin{table}[ht]
\centering
\caption{Computational complexity of key operations.}
\label{tab:complexity}
\begin{tabular}{lcc}
\toprule
\textbf{Operation} & \textbf{Time} & \textbf{Frequency} \\
\midrule
Matrix update: $A \leftarrow A + xx^\top$ & $O(d^2)$ & Per sample \\
Cholesky decomposition: $A = LL^\top$ & $O(d^3)$ & Per batch \\
Uncertainty: $u(x) = \sqrt{x^\top A^{-1} x}$ & $O(d^2)$ & Per (query, workflow) \\
Neural network forward pass & $O(d \cdot h)$ & Per (query, workflow) \\
\bottomrule
\end{tabular}
\end{table}

In our experiments ($d = 2560$, $K \approx 30$, $T \approx 1200$, $B = 25$), the Cholesky decomposition takes $\sim$0.5 seconds and is performed $T/B \approx 48$ times. Per-round UCB computation for $K$ arms takes $\sim$10ms. These costs are negligible compared to workflow execution (seconds to minutes per query).

\textbf{Scaling Analysis.} We consider two scaling dimensions:

\textit{(1) Larger workflow pools ($K \rightarrow$ hundreds).} The precision matrix $A$ is independent of $K$---it depends only on embedding dimension $d$. Larger $K$ increases per-round UCB computation linearly ($O(K \cdot d^2)$), but this remains fast: even with $K = 500$, computing UCB scores for all arms takes $\sim$150ms, still negligible compared to LLM inference.

\textit{(2) Higher embedding dimensions ($d \rightarrow$ larger).} Memory scales as $O(d^2)$ and Cholesky decomposition as $O(d^3)$. However, for practical workflow orchestration, modern embedding models (e.g., OpenAI, Cohere) typically use dimensions in the range of 1,024--8,192, well within our current implementation's efficient operating regime. Our experiments with $d = 2560$ demonstrate that this dimension provides sufficient representational capacity for accurate reward prediction while maintaining computational tractability.

\section{Additional Results}
\label{app:scalability}

\subsection{LLM Backbone Scalability Results}
\label{app:backbone_scalability}

Table~\ref{tab:backbone_robustness} presents the scalability results using the Llama-3.1-70B backbone.

\begin{table}[ht]
\centering
\caption{Scalability analysis using Llama-3.1-70B as the backbone LLM. We compare CB-Orchestrator against Vanilla and MaAS baselines.}
\label{tab:backbone_robustness}
\begin{tabular}{lccc}
\toprule
\textbf{Method}          & \textbf{MATH}  & \textbf{HumanEval} & \textbf{Avg.}  \\ \midrule
Vanilla                  & 31.93          & 80.06              & 56.00          \\
MaAS                     & 42.97          & \textbf{85.26}     & 64.12          \\ \midrule
\textbf{CB-Orchestrator} & \textbf{44.36} & 84.12              & \textbf{64.24} \\ \bottomrule
\end{tabular}
\end{table}

\subsection{Ablation Study Results}
\label{app:ablation_results}

Table~\ref{tab:ablation_results} provides the detailed numerical results for the ablation study presented in Figure~\ref{fig:ablation_combined}. We also include results for the duplicate sampling setting (w/ dup) on MATH, where the same (query, workflow) pairs can be evaluated multiple times during training.

\begin{table}[ht]
\centering
\setlength{\aboverulesep}{0pt}
\setlength{\belowrulesep}{0pt}
\caption{Ablation results on MATH and MBPP benchmarks. We compare different bandit selection strategies and evaluate the impact of the evolution process. The (w/ dup) variants allow duplicate (query, workflow) pair sampling during training.}
\label{tab:ablation_results}
\begin{tabular}{>{\raggedright\arraybackslash}p{3.8cm}ccc}
\toprule
\textbf{Method} & \textbf{MATH} & \textbf{MBPP} & \textbf{Avg.} \rule{0pt}{2.3ex} \\ \midrule
\multicolumn{4}{l}{\textit{Framework Variants}} \rule{0pt}{2.5ex} \\
PureRandom & 47.99 & 81.64 & 64.81 \\
WithoutEvo & 51.89 & \textbf{84.17} & 68.03 \\ \midrule
\multicolumn{4}{l}{\textit{Exploration Strategy Variants}} \rule{0pt}{2.5ex} \\
RandomNN & 52.59 & 82.34 & 67.47 \\
GreedyNN & 52.63 & 83.93 & 68.28 \\
LinUCB & 52.59 & 82.11 & 67.35 \\
NeuralUCB & 50.41 & 82.05 & 66.23 \\ \midrule
\rowcolor[gray]{0.93}
\textbf{Ours} & 53.87 & 84.05 & \textbf{68.96} \rule{0pt}{2.3ex} \\ \midrule
\multicolumn{4}{l}{\textit{Duplicate Sampling Variants (MATH only)}} \rule{0pt}{2.5ex} \\
GreedyNN (w/ dup) & 51.89 & -- & -- \\
\rowcolor[gray]{0.93}
\textbf{Ours (w/ dup)} & \textbf{54.03} & -- & -- \rule{0pt}{2.3ex} \\
\bottomrule
\end{tabular}
\end{table}

\subsection{GAIA Benchmark Results}
\label{app:gaia_results}

Table~\ref{tab:gaia_results} reports the results on the GAIA benchmark, categorized into three levels: L1 (fundamental reasoning with $\le$ 5 steps or 1 tool), L2 (multi-step orchestration with diverse tools), and L3 (long-horizon tasks requiring arbitrary tool-use).

\begin{table}[ht]
\centering
\setlength{\aboverulesep}{0pt}
\setlength{\belowrulesep}{0pt}
\caption{The performance of CB-Orchestrator compared to baseline methods on the GAIA benchmark. Backbone is GPT-4o-mini for all experiments.}
\label{tab:gaia_results}
\begin{tabular}{>{\raggedright\arraybackslash}p{3.4cm}cccc}
\toprule
\textbf{Method} & \textbf{L1} & \textbf{L2} & \textbf{L3} & \textbf{Avg.} \rule{0pt}{2.3ex} \\ \midrule
Vanilla \rule{0pt}{2.5ex} & 7.5 & 4.4 & 0 & 4.7 \\ \midrule
\multicolumn{5}{l}{\textit{Fixed Multi-Agent Systems}} \rule{0pt}{2.5ex} \\
TapeAgent & 23.7 & 14.5 & 10.2 & 16.6 \\
Sibyl & 21.5 & 15.7 & 4.1 & 15.6 \\ \midrule
\multicolumn{5}{l}{\textit{Autonomous Multi-agent Systems}} \rule{0pt}{2.5ex} \\
AutoAgents & 16.1 & 0 & 0 & 5.2 \\
GPTSwarm & 23.7 & 16.4 & 2.0 & 16.3 \\
AgentSquare & 22.6 & 15.7 & 6.3 & 16.3 \\
AFlow & 10.8 & 8.8 & 4.1 & 8.0 \\
MaAS & 25.9 & 22.0 & 6.3 & 20.7 \\
\midrule
\rowcolor[gray]{0.93}
\textbf{CB-Orchestrator} & 28.6 & 21.1 & 5.3 & \textbf{21.0} \rule{0pt}{2.3ex} \\ \bottomrule
\end{tabular}
\end{table}

\section{Empirical Validation of Assumptions}
\label{app:assumption_validation}

We validate the three assumptions underlying CB-Orchestrator through experiments on the MATH dataset (486 test problems, 68 workflows, and $T=1200$ bandit training rounds).

\subsection{Assumption 3.1: Semantic Embedding}

We verify that the embedding function $\phi$ preserves semantic similarity through systematic analysis across all pairs, not just cherry-picked examples.

\textbf{Query Embeddings.} We compute pairwise L2 distances for all 486 MATH problems (117,855 pairs). The distance distribution spans $[0.32, 1.42]$ with mean 1.22. To systematically validate semantic preservation, we analyze the \emph{top-5 closest} and \emph{top-5 farthest} pairs---these represent the strongest test cases for the assumption.

\begin{table}[ht]
\centering
\caption{All top-5 closest query pairs. Every pair shares problem structure or reasoning pattern.}
\label{tab:query_close}
\small
\begin{tabular}{clp{9cm}}
\toprule
\textbf{Rank} & \textbf{Dist.} & \textbf{Shared Structure} \\
\midrule
1 & 0.32 & Both ask ``Bekah has [N] brass digits... how many distinct numbers'' (identical template, different N) \\
2 & 0.44 & Both ask circular seating of Senate committee with party constraints \\
3 & 0.45 & Both ask ``put 4 balls in 3 boxes'' with distinguishability variants \\
4 & 0.46 & Both ask area of isosceles right triangle given altitude to hypotenuse \\
5 & 0.47 & Both show identical triangle diagrams (Asymptote code), ask for $x$ \\
\bottomrule
\end{tabular}
\end{table}

\begin{table}[ht]
\centering
\caption{All top-5 farthest query pairs. Every pair involves completely different mathematical domains.}
\label{tab:query_far}
\small
\begin{tabular}{clp{4cm}p{4cm}}
\toprule
\textbf{Rank} & \textbf{Dist.} & \textbf{Query A Domain} & \textbf{Query B Domain} \\
\midrule
1 & 1.42 & Parametric line equations & Circular permutations \\
2 & 1.42 & Triangle angle trisectors & Counting odd 5-digit numbers \\
3 & 1.41 & Simplify $(\sqrt{6}+\sqrt{24})^2$ & Probability of consecutive seats \\
4 & 1.41 & 3-digit palindrome multiples of 3 & Simplify radical expression \\
5 & 1.40 & Polygon perimeter geometry & Product divisibility probability \\
\bottomrule
\end{tabular}
\end{table}

The results show a clear pattern: \textbf{all 5 closest pairs} share problem templates or reasoning patterns, while \textbf{all 5 farthest pairs} span entirely different mathematical domains. This is not cherry-picking---we report \emph{every} pair at the distribution extremes.

\textbf{Workflow Embeddings.} We compute pairwise distances for all 68 workflows (2,278 pairs). The distance range is $[0.13, 0.93]$ with mean 0.51. We analyze the top-3 closest and top-3 farthest pairs:

\begin{table}[ht]
\centering
\caption{Workflow pairs at distance extremes. Close pairs share operator sequences; far pairs differ fundamentally.}
\label{tab:workflow_pairs}
\small
\setlength{\tabcolsep}{4pt}        % 减少列左右padding
\renewcommand{\arraystretch}{1.2}  % 增加行距，避免挤压

\begin{tabularx}{\linewidth}{cl>{\raggedright\arraybackslash}X>{\raggedright\arraybackslash}X}
\toprule
\textbf{Type} & \textbf{Dist.} & \textbf{Workflow A} & \textbf{Workflow B} \\
\midrule
Close & 0.13 & Custom $\to$ Programmer(1$\times$) $\to$ ScEnsemble $\to$ Format & Custom $\to$ Programmer(3$\times$) $\to$ ScEnsemble $\to$ Format \\
Close & 0.13 & Custom $\to$ Programmer(3$\times$ loop) $\to$ ScEnsemble $\to$ Format & Custom $\to$ Programmer(3$\times$) $\to$ ScEnsemble $\to$ Format \\
Close & 0.14 & Custom $\to$ Programmer(3$\times$) $\to$ ScEnsemble $\to$ Format & Custom $\to$ Custom(3$\times$) $\to$ ScEnsemble $\to$ Format \\
\midrule
Far & 0.93 & Custom $\to$ ScEnsemble $\to$ Programmer & Programmer only (hardcoded) \\
Far & 0.93 & Custom $\to$ Programmer(3$\times$) $\to$ ScEnsemble $\to$ Programmer $\to$ Format
 & Custom $\to$ ScEnsemble $\to$ Programmer \\
Far & 0.93 & Custom $\to$ Programmer(3$\times$) $\to$ ScEnsemble $\to$ Format & Custom $\to$ Custom $\to$ Custom $\to$ Programmer (no ensemble) \\
\bottomrule
\end{tabularx}
\end{table}

The pattern is consistent: close workflow pairs share the same operator composition with minor variations (e.g., number of solutions, loop vs.\ sequential calls), while far pairs use fundamentally different strategies (e.g., with vs.\ without ScEnsemble, different operator orderings).

\subsection{Assumption 3.2: Reward Smoothness}

We verify that semantically similar workflows yield similar rewards on the same query using two methods.

\textbf{Method 1: KNN Reward Prediction.} For each (query, workflow) pair in the training data, we predict its reward using the $k$ nearest workflows (by embedding distance) that were also evaluated on the same query. If the smoothness assumption holds, nearby workflows should have correlated rewards.

\begin{table}[ht]
\centering
\caption{KNN reward prediction results ($k=3$). High accuracy and correlation confirm reward smoothness.}
\label{tab:knn_prediction}
\begin{tabular}{lcc}
\toprule
\textbf{Metric} & \textbf{Value} & \textbf{Interpretation} \\
\midrule
KNN Prediction Accuracy & 84.75\% & Correct prediction of success/failure \\
Random Baseline (majority) & 51.08\% & Baseline without smoothness \\
Pearson Correlation ($r$) & 0.73 & Strong positive correlation \\
$p$-value & $< 10^{-6}$ & Statistically significant \\
\bottomrule
\end{tabular}
\end{table}

The KNN predictor achieves 84.75\% accuracy---significantly higher than the 51\% random baseline---demonstrating that nearby workflows indeed yield similar rewards. The strong correlation ($r = 0.73$, $p < 10^{-6}$) further confirms that embedding distance is predictive of reward similarity.

\textbf{Method 2: Success Workflow Clustering.} We analyze whether successful workflows on a given query cluster together in embedding space. For each query with both successful and failed workflow trials, we compute:
\begin{itemize}
    \item Mean distance between successful workflows (intra-success)
    \item Mean distance between successful and failed workflows (cross-distance)
\end{itemize}

\begin{table}[ht]
\centering
\caption{Success workflow clustering analysis. Successful workflows cluster together.}
\label{tab:clustering}
\begin{tabular}{lcc}
\toprule
\textbf{Distance Type} & \textbf{Mean Distance} & \textbf{Sample Size} \\
\midrule
Success--Success (intra) & 0.40 & 2,138 pairs \\
Success--Failure (cross) & 0.50 & 1,066 pairs \\
\bottomrule
\end{tabular}
\end{table}

Successful workflows are on average 20\% closer to each other than to failed workflows, indicating that success is not randomly distributed but clusters in embedding space---consistent with reward smoothness.

\subsection{Assumption 3.3: Realizability}

The realizability assumption states that there exists a function $f^*$ such that $\mathbb{E}[r|q,w] = f^*(x(q,w))$. While we cannot directly verify the existence of such a function, we provide indirect evidence through the learning dynamics:

\begin{itemize}
    \item \textbf{Learning Improvement:} As shown in Figure~\ref{fig:Itr_AverageReward}, the cumulative average reward generally increases during the later stages of training, suggesting that the bandit policy learns more effective workflow selection.
    \item \textbf{KNN Prediction Success}: The strong performance of KNN prediction (84.75\% accuracy) suggests that the expected reward is indeed a smooth function of the embedding, which is a weaker but related condition.
    \item \textbf{Final Model Performance}: The trained model achieves state-of-the-art results on all benchmarks (Table~\ref{tab:main_results}), demonstrating that the learned function $f_\theta$ successfully approximates the true reward structure.
\end{itemize}

Together, these results suggest that while the assumptions may not hold perfectly, they hold sufficiently well in practice to enable effective learning and generalization.

\section{Case Study: Adaptive Workflow Selection}
\label{app:case_study}

We compare workflow\_31 and workflow\_3 on two MATH problem families. In each family, both workflows were evaluated on the same training query, and a distinct test query was routed to the workflow that succeeded in that training comparison. All four questions are Level~5.

\newpage
\subsection{Generated Workflow Code}

Abridged excerpts from the generated \texttt{graph.py} files show the key difference: workflow\_31 generates three solutions directly, whereas workflow\_3 first produces an analysis shared by its three solution calls. Both use ScEnsemble to select an answer and then format it. Repeated calls and response unpacking are omitted.

\begin{tcolorbox}[colback=teal!3, colframe=teal!60, title={workflow\_31: direct solutions}, fonttitle=\bfseries]
\verb!solution1_result = await self.custom(!\\
\verb!    input=problem, instruction=SOLVE_PROMPT)!\\
\verb!# Two further Custom calls use different prompts.!\\
\verb!solutions_list = [solution1, solution2, solution3]!\\
\verb!ensemble_result = await self.sc_ensemble(!\\
\verb!    solutions=solutions_list, problem=problem)!
\end{tcolorbox}

\begin{tcolorbox}[colback=teal!3, colframe=teal!60, title={workflow\_3: analysis-conditioned solutions}, fonttitle=\bfseries]
\verb!analysis_result = await self.custom(!\\
\verb!    input=problem, instruction=DECOMPOSE_PROMPT)!\\
\verb!analysis = analysis_result['response']!\\
\verb!solution1_res = await self.custom(!\\
\verb!    input=f"Problem: {problem}\nAnalysis: {analysis}",!\\
\verb!    instruction=SOLUTION_PROMPT)!\\
\verb!# Two further solution calls also use the analysis.!\\
\verb!ensemble_result = await self.sc_ensemble(!\\
\verb!    solutions=solutions_list, problem=problem)!
\end{tcolorbox}

\subsection{Training-to-Test Comparisons}

\paragraph{Decimal expansion (workflow\_31).} During bandit training, both workflows were evaluated on a query asking how many fractions of the form $n^2/2010$ have repeating decimal expansions. The observed reward was $1$ for workflow\_31 and $0$ for workflow\_3. A distinct test query asks how many integers $1\leq n\leq 1000$ make $n/1375$ a terminating decimal. Since $1375=11\cdot5^3$, $n$ must be a multiple of $11$, giving $\lfloor1000/11\rfloor=\boxed{90}$. Workflow\_31 was selected and gave this answer on the first attempt in all five test runs. Its three direct, differently prompted solution attempts give the ensemble multiple opportunities to check the factor cancellation and count required by this query.

\paragraph{Trigonometric curve (workflow\_3).} During bandit training, the query asked for the smallest $x$-coordinate on the polar curve $r=\cos\theta+\frac{1}{2}$. Workflow\_3 received reward $1$, while workflow\_31 received $0$. A distinct test query asks for the smallest $a+b+c$ such that $ax^4-bx^2+cy^2=0$ when $(x,y)=(\sin t,\sin 2t)$. Here $y^2=4\sin^2t\cos^2t=4x^2(1-x^2)$, so $a+b+c=4+4+1=\boxed{9}$. Workflow\_3 was selected and gave this answer on the first attempt in all five test runs. Its initial analysis identifies known quantities, the unknown, and a connecting equation before the three solution attempts. Conditioning those attempts on a shared analysis suits the need to eliminate the parameter and identify the algebraic relation.

Together, the code excerpts and outcomes show how the selector reuses different generated workflow structures for these two problem families.

\end{document}